\documentclass[10pt,twocolumn,letterpaper]{article}

\usepackage[pagenumbers]{cvpr} 

\usepackage[dvipsnames]{xcolor}

\definecolor{cvprblue}{rgb}{0.21,0.49,0.74}
\usepackage[pagebackref,breaklinks,colorlinks,citecolor=cvprblue]{hyperref}

\usepackage[accsupp]{axessibility} 

\usepackage{url}            
\usepackage{booktabs}       
\usepackage{amsfonts}       
\usepackage{nicefrac}       
\usepackage{microtype}      
\usepackage{makecell}       
\usepackage{multicol} 
\usepackage{multirow}
\usepackage{xspace}
\usepackage{multirow}
\usepackage{adjustbox}
\usepackage{colortbl}
\usepackage{duckuments}
\usepackage{lipsum}
\usepackage{enumitem}
\usepackage{kotex}
\usepackage{tikz}
\usepackage{amsmath}
\usepackage{subcaption}
\usepackage{graphicx}
\usepackage{tabularx}
\newcolumntype{Y}{>{\centering\arraybackslash}X}
\usepackage{rotating}
\usepackage{color}
\usepackage[utf8]{inputenc}

\usepackage{amsthm}

\newtheorem{theorem}{Theorem}[section]

\theoremstyle{definition}
\newtheorem{definition}{Definition}[section]

\theoremstyle{remark}

\usepackage{algorithm}

\usepackage[noend]{algpseudocode}
\usepackage{float}
\usepackage{amsmath}
\usepackage{commath}
\algrenewcommand\algorithmicrequire{\textbf{Input:}}

\algrenewcommand\algorithmicensure{\textbf{Output:}}
\usepackage{flexisym}
\usepackage{relsize}\usepackage{amssymb}
\usepackage{wrapfig}

\definecolor{BrickRed}{rgb}{0.6,0,0}
\definecolor{RoyalBlue}{rgb}{0,0,0.8}
\definecolor{Tdgreen}{rgb}{0,0.4,0.7}

\DeclareMathOperator*{\argmax}{arg\,max}

\usepackage{subcaption} 
\usepackage{bbold} 

\newcommand{\std}{0.7}

\newcommand{\system}{AETTA}

\def\paperID{3695} 
\def\confName{CVPR}
\def\confYear{2024}

\title{\system{}: Label-Free Accuracy Estimation for Test-Time Adaptation}

\makeatletter
\newcommand{\specificthanks}[1]{\@fnsymbol{#1}}
\newcommand{\kaist}{\textsuperscript{\specificthanks{2}}}
\newcommand{\nokia}{\textsuperscript{\specificthanks{3}}}

\author{%
Taeckyung Lee\kaist \quad Sorn Chottananurak\kaist \quad Taesik Gong\nokia \quad Sung-Ju Lee\kaist
\vspace{2mm} \\
\kaist KAIST \quad \nokia Nokia Bell Labs
\vspace{0.5mm} \\
{\tt \small \{taeckyung,sorn111930,profsj\}@kaist.ac.kr, \tt \small taesik.gong@nokia-bell-labs.com}\\
}

\begin{document}
\maketitle

\begin{abstract}

Test-time adaptation (TTA) has emerged as a viable solution to adapt pre-trained models to domain shifts using unlabeled test data. However, TTA faces challenges of adaptation failures due to its reliance on blind adaptation to unknown test samples in dynamic scenarios. Traditional methods for out-of-distribution performance estimation are limited by unrealistic assumptions in the TTA context, such as requiring labeled data or re-training models. 
To address this issue, we propose \system{}, a label-free accuracy estimation algorithm for TTA. 
We propose the prediction disagreement as the accuracy estimate, calculated by comparing the target model prediction with dropout inferences.
We then improve the prediction disagreement to extend the applicability of \system{} under adaptation failures. Our extensive evaluation with four baselines and six TTA methods demonstrates that \system{} shows an average of 19.8\%p more accurate estimation compared with the baselines.
We further demonstrate the effectiveness of accuracy estimation with a model recovery case study, showcasing the practicality of our model recovery based on accuracy estimation. The source code is available at \url{https://github.com/taeckyung/AETTA}.

\end{abstract}
 
\section{Introduction}\label{sec:intro}

The rise of deep learning has impacted various fields with remarkable achievements~\cite{7780459, ronneberger2015u, 10.1145/3422622, vaswani2017attention, brown2020language}. In real-world deep learning applications, the divergence between training and test data, known as domain shifts,  often leads to poor accuracy. 
For instance, object detection models encountering previously unseen data (\eg, variations of objects) or distributional shifts (\eg, weather changes) might suffer from performance degradation. To overcome this challenge, Test-Time Adaptation (TTA)~\cite{note, sar, lame, cotta, rotta, tent, sotta, eata} has been regarded as a promising solution recently and actively studied. TTA aims to adapt pre-trained models to domain shifts on the fly with only unlabeled test data.

Despite recent advances in TTA, significant challenges hinder its practical applications. The core issue is that TTA's reliance on unlabeled test-domain samples makes TTA susceptible to adaptation failures, especially in dynamic environments where the domain continuously changes~\cite{sar, rdumb}. Although recent TTA studies deal with dynamic test streams in TTA~\cite{note, sar, cotta, rotta, sotta}, the inherent risk of TTA--blind adaptation to unseen test samples without ground-truth labels--remains a critical vulnerability. Notably, the absence of ground-truth labels makes it difficult to monitor the correctness of the adaptation. While various out-of-distribution performance estimation approaches have been proposed~\cite{agreement_on_the_line, accuracy_on_the_line, chen2021ensemble, DoC}, such methods necessitate labeled train data for accuracy estimation, which is impractical for TTA scenarios.

\begin{figure}[t]
    \centering
    \includegraphics[width=1\linewidth]{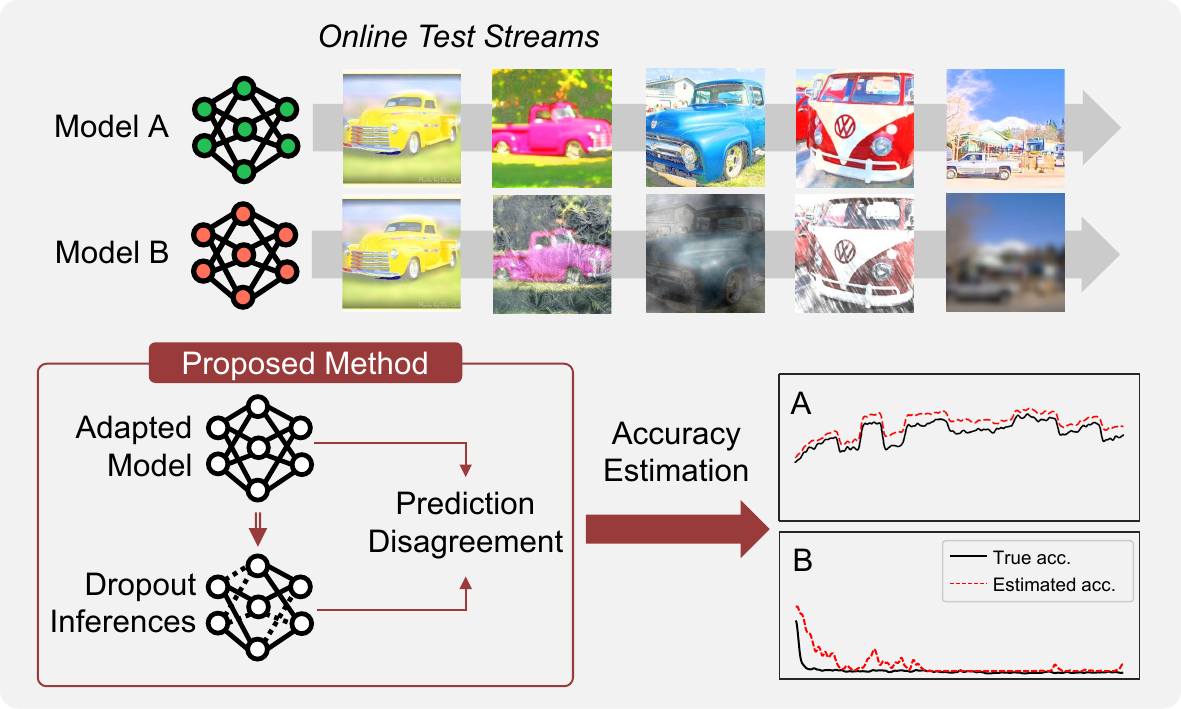}
    \caption{\system{} estimates the model's accuracy after adaptation using unlabeled test data without needing source data or ground-truth labels. \system{} can be integrated into existing TTA methods to estimate their accuracy under various scenarios.
    }\label{fig:concept}
\end{figure}

In light of these challenges, we propose \system{} (Accuracy Estimation for Test-Time Adaptation), a novel accuracy estimation method designed for TTA without reliance on labeled data or source data access (Figure~\ref{fig:concept}). 
\system{} leverages \emph{prediction disagreement with dropout inferences}, where the prediction disagreement between the adapted model and dropout inferences serves as a basis for performance estimation. To enhance \system{}'s robustness to adaptation failure scenarios, we propose \textit{robust disagreement equality} that dynamically adjust the accuracy estimates based on model failures. The key idea is to extend the well-calibration assumption (\ie, predicted probabilities of expected model predictions are neither over-/under-confident~\cite{disagreement}) to cover over-confident models (\eg, adaptation failures) via adaptive scaling of the predicted probability.
In addition, we provide theoretical analysis on how \system{} can estimate accuracy with unlabeled test data. 

We evaluate \system{} on three TTA benchmarks (CIFAR10-C, CIFAR100-C, and ImageNet-C~\cite{cifarc}) with two scenarios of fully TTA (\ie, adapting to each corruption)~\cite{tent} and continual TTA (\ie, continuously adapting to 15 corruptions)~\cite{cotta}. We evaluate the accuracy estimation of \system{} integrated with six state-of-the-art TTA algorithms~\cite{tent, eata, sar, cotta, rotta, sotta}. We compare \system{} with four baselines that could be applied in the TTA setting. The result illustrates that \system{} shows an average of 19.8\%p more accurate estimation compared with the baselines in various TTA methods and evaluation scenarios.

Furthermore, we explore the impact of performance estimation in TTA through a case study where we avoided undesirable accuracy drops in TTA based on \system{}. We propose a simple model recovery algorithm, 
which resets the model when consecutive estimated accuracy degradation or sudden accuracy drop are observed.
Our case study shows that our model recovery algorithm with accuracy estimation achieved 11.7\%p performance improvement, outperforming the best baseline that knows when distribution changes by 3.0\%p.
The result shows an example where accuracy estimation could benefit TTA in practice. 

\section{Preliminaries}\label{sec:background}

\subsection{Test-Time Adaptation (TTA)} Consider the source data distribution $\mathcal{D^{\mathcal{S}}}$, and the target data distribution $\mathcal{D^{\mathcal{T}}}$ and its random variable ${({X}, {Y})}$, where $Y$ is typically unknown to the learning algorithm, and $K$ is total number of classes. The covariate shift assumption~\cite{quinonero2008dataset} asserts a disparity between the source and target data distributions, defined by $\mathcal{D}^{\mathcal{S}}(\mathbf{x}) \neq \mathcal{D}^{\mathcal{T}}(\mathbf{x})$ while maintaining consistency in the conditional label distribution: $\mathcal{D}^{\mathcal{S}}(y|\mathbf{x}) = \mathcal{D}^{\mathcal{T}}(y|\mathbf{x})$.

Let $h \sim {\mathcal{H}}_{\mathcal{A}}$ denote a hypothesis that predicts a single class for a single input and $f$ denote a corresponding softmax value before class prediction. We define the hypothesis space ${\mathcal{H}}_{\mathcal{A}}$ as a hypothesis space ${\mathcal{H}}$ induced by a stochastic training algorithm $\mathcal{A}$~\cite{disagreement}. The stochasticity could arise from a different random initialization or data ordering.

Assuming an off-the-shelf model $h_0 \sim {\mathcal{H}}_{\mathcal{A}}$ pre-trained on $\mathcal{D^{\mathcal{S}}}$, the goal of (fully) test-time adaptation (TTA)~\cite{tent} is to adapt $h_0$ for the target distribution $D^{\mathcal{T}}$ to produce $h$, using a batch of the unlabeled test set in an online manner.

\subsection{Accuracy Estimation in TTA}

We adopt a common TTA setup where source data is unavailable and target test data lacks labels~\cite{tent, eata, sar, cotta, rotta, sotta}.
The objective of TTA accuracy estimation is to predict the test accuracy (or error) with unlabeled test streams.

Given an adapted model $h(\cdot; \Theta)$ at time $t$, 
we denote the test error of model $h(\cdot; \Theta)$ by:
\begin{equation}
    {\tt Err}_{\mathcal{D}^{\mathcal{T}}} (h)  \triangleq  \mathbb{E}_{\mathcal{D}^{\mathcal{T}}} [ \mathbb{1} ( h({X}) \neq Y ) ].
\end{equation}
Note that we use the terms test accuracy and test error depending on the context, and the sum of them is 1.
Given the temporal nature of TTA, we consider estimating the accuracy of the model $h(\cdot; \Theta)$--which has been updated before time $t$--with the test batch $\mathbf{X}_t$. Following the estimation, the test batch $\mathbf{X}_t$ is used for adaptation.

\section{Methodology}

\subsection{Disagreement Equality}\label{sec:disagreement-equality}

We introduce an approach for estimating the test error of a model that is adapted at test time. The key idea is to compare the model's output against outputs generated through dropout inference. Remarkably, this estimation process does not rely on access to the original training or labeled test data, which contrasts with existing accuracy estimation methods~\cite{agreement_on_the_line, disagreement, accuracy_on_the_line, chen2021ensemble, DoC}.
For example, generalization disagreement equality (GDE)~\cite{disagreement} proposes a theoretical ground for estimating model error by measuring the disagreement rate between two networks. However, GDE requires multiple pre-trained models from different training procedures to calculate the disagreement rate.

Instead of multiple pre-trained models, our strategy utilizes dropout inference sampling, a technique where random parts of a model's intermediate layer outputs are omitted during the inference process~\cite{dropout}. From a single adapted model, we simulate the behavior of independent and identically distributed (i.i.d.) models by dropout inference sampling.

\begin{definition}\label{def:efa}
    The hypothesis space ${\mathcal{H}}_{\mathcal{A}}$ satisfies the \textbf{dropout independence} if for any $h \sim {\mathcal{H}}_{\mathcal{A}}$, $h$ and its dropout inference samples are i.i.d. over ${\mathcal{H}}_{\mathcal{A}}$.
\end{definition}

To estimate the accuracy of the model, we propose \textbf{prediction disagreement with dropout inferences (PDD)} that calculates a disagreement between the adapted model $h(\cdot; \Theta)$ and the dropout inferences $h(\cdot; \Theta^{{\tt dropout}})$ with respect to test samples as:
\begin{equation}
\begin{aligned}
    {\tt PDD}_{\mathcal{D}^{\mathcal{T}}} \! (h) \triangleq \mathbb{E}_{\mathcal{D}^{\mathcal{T}}} \!\!\! \left[ \frac{1}{N} \! \sum_{i=1}^N \mathbb{1} \! \left[ h({X} ; \Theta) \neq {h}({X} ; \Theta^{{\tt dropout}_i})  \right] \! \right] \!\!,
\end{aligned}
\end{equation}
where $N$ is the number of dropout inferences.

We now provide the theoretical background to estimate test error with PDD. 
We first define the expectation function $\Tilde{h}$~\cite{disagreement} over hypothesis space ${\mathcal{H}}_{\mathcal{A}}$, which produces probability vector of size $K$. For $k$-th element $\Tilde{h}_k(\mathbf{x})$, we define:
\begin{equation}
    \Tilde{h}_k(\mathbf{x})  \triangleq \mathbb{E}_{h \sim {\mathcal{H}}_{\mathcal{A}}} [ \mathbb{1} [ h(\mathbf{x}) = k]],
\end{equation}
which indicates the probability of a sample $\mathbf{x}$ sampled from ${D^{\mathcal{T}}}$ being classed as the class $k$. 
Note that the expectation function does not represent the model's accuracy; it indicates the probability of the input being classified as a particular class, regardless of the ground truth labels.

Then, we define a confidence-prediction calibration assumption, indicating that the value of $\Tilde{h}$ for a particular class equals the probability of the sample having the same ground-truth label~\cite{disagreement}.

\begin{definition}\label{def:cpc}
    The hypothesis space ${\mathcal{H}}_{\mathcal{A}}$ and corresponding expectation function $\Tilde{h}$ satisfies \textbf{confidence-prediction calibration}\footnote{We rename the term from class-wise calibration~\cite{disagreement} to clearly state the purpose of the calibration.} on $\mathcal{D}^{\mathcal{T}}$ if for any confidence value $q \in [0, 1]$ and class $k \in [1, \cdots ,K]$:
    \begin{equation}
        p({Y} = k | \Tilde{h}_k ({X}) = q) = q.
    \end{equation}
\end{definition}

With PDD and the assumption of dropout independence and confidence-prediction calibration, we are able to estimate the model's prediction error $h$~(Theorem~\ref{thm:class-wise}). Detailed proof is provided in the Appendix~\ref{sec:app-class-wise}. 

\begin{theorem}[Disagreement Equality]\label{thm:class-wise}
    If the hypothesis space ${\mathcal{H}}_{\mathcal{A}}$ and corresponding expectation function $\Tilde{h}$ satisfies dropout independence and confidence-prediction calibration, prediction disagreement with dropouts (PDD) approximates the test error over ${\mathcal{H}}_{\mathcal{A}}$:
    \begin{equation}
        \mathbb{E}_{h \sim {\mathcal{H}}_{\mathcal{A}}} [{\tt{Err}}_{\mathcal{D}^{\mathcal{T}}}(h)] =
        \mathbb{E}_{h \sim {\mathcal{H}}_{\mathcal{A}}} [{\tt{PDD}}_{\mathcal{D}^{\mathcal{T}}}(h)].
    \end{equation}
\end{theorem}

\subsection{Robust Disagreement Equality}\label{sec:robust-disagreement-equality}

\begin{figure}[t]
    \centering
    \begin{subfigure}[t]{0.55\linewidth}
    \centering
    \includegraphics[width=0.95\linewidth]{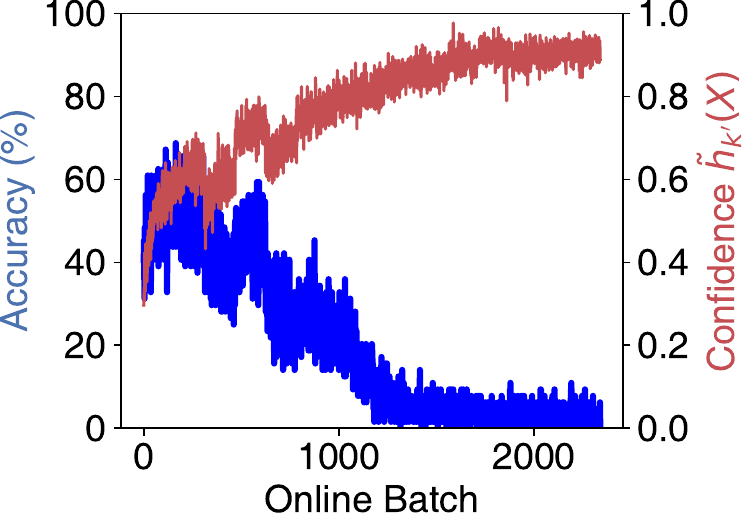}
    \caption{Test batch accuracy and confidence.}
    \end{subfigure}
    \begin{subfigure}[t]{0.44\linewidth}
    \centering
    \includegraphics[width=0.95\linewidth]{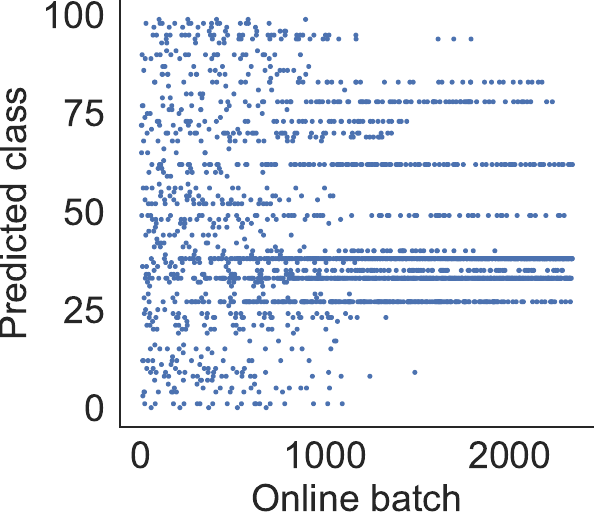}
    \caption{Predicted class distribution.}
    \label{fig:acc_conf_dist}
    \end{subfigure}
    \caption{Batch-wise accuracy, confidence, and prediction distribution when a model failed to adapt.  TENT~\cite{tent} is used on CIFAR100-C with continually changing domains. The model becomes over-confident, and predictions are skewed.
    }\label{fig:acc_conf}
\end{figure}

\begin{figure*}[t]
  \centering
    \begin{subfigure}[t]{0.16\linewidth}
    \centering
    \includegraphics[width=1\linewidth]{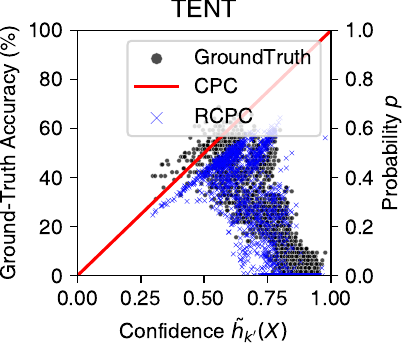}
    \end{subfigure}
    \begin{subfigure}[t]{0.16\linewidth}
    \centering
    \includegraphics[width=1\linewidth]{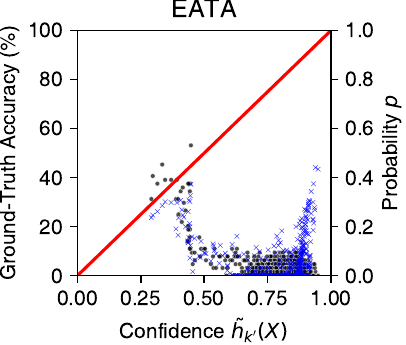}
    \end{subfigure}
    \begin{subfigure}[t]{0.16\linewidth}
    \centering
    \includegraphics[width=1\linewidth]{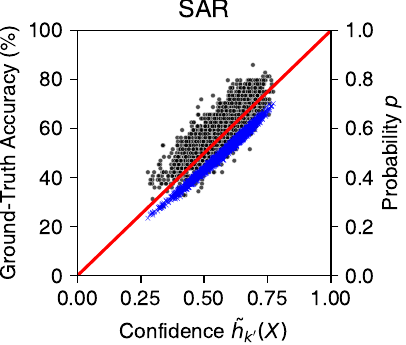}
    \end{subfigure}
    \begin{subfigure}[t]{0.16\linewidth}
    \centering
    \includegraphics[width=1\linewidth]{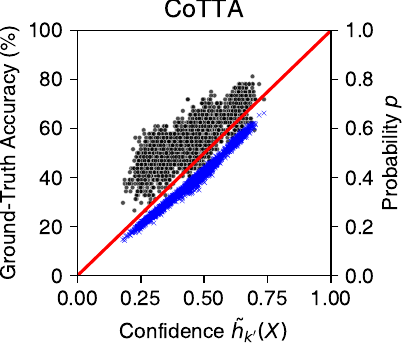}
    \end{subfigure}
    \begin{subfigure}[t]{0.16\linewidth}
    \centering
    \includegraphics[width=1\linewidth]{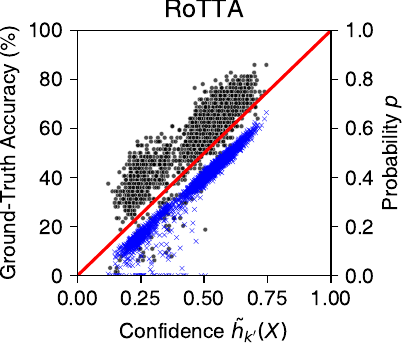}
    \end{subfigure}
    \begin{subfigure}[t]{0.16\linewidth}
    \centering
    \includegraphics[width=1\linewidth]{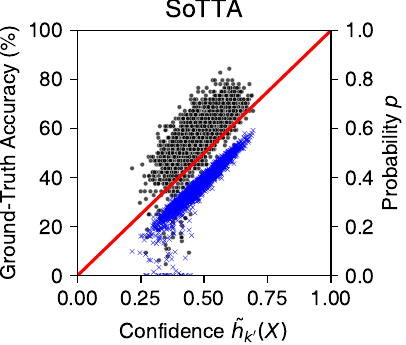}
    \end{subfigure}
  \caption{Correlations between the confidence value of estimated expectation function $\Tilde{h}$ and (1) ground-truth accuracy (GroundTruth), (2) conditional probability $p(Y = k' | \Tilde{h}_{k'} (X) = q)$ of confidence-prediction calibration (CPC), and (3) robust confidence-prediction calibration (RCPC). We used six TTA methods in CIFAR100-C with continual domain changes. We observed accuracy degradation in TENT and EATA and improvement in SAR, CoTTA, RoTTA, and SoTTA. When models failed to adapt, the original CPC misaligned with the ground truth. In contrast, our WCPC dynamically scaled the probability $p$, thus showing better alignment.
  }\label{fig:acc_corr}
\end{figure*}

Adaptation failures in TTA are often coupled with over-confident incorrect predictions. Figure~\ref{fig:acc_conf} shows an illustrative example of this case; as the expectation function's accuracy drops, the confidence increases, and predictions are skewed towards a few classes\footnote{\label{footnote:expectation-function}Using the probabilistic property of expectation of dropout inferences~\cite{dropout}, we approximate $\Tilde{h}(X)$ as $\mathbb{E}_{{\mathcal{H}}_{\mathcal{A}}} [ \mathbb{E}_{{\tt dropout}} [h(X; \Theta^{\tt dropout})] ]$.}.
This violates the confidence-prediction calibration, leading to a high misalignment between test error and PDD (red lines in Figure~\ref{fig:acc_corr}).

To tackle the issue, we propose a \textbf{robust confidence-prediction calibration} to provide the theoretical ground of accuracy estimation for both well-calibrated or over-confident expectation function $\Tilde{h}$.

\begin{definition}\label{def:scpc}
    The hypothesis space ${\mathcal{H}}_{\mathcal{A}}$ and corresponding expectation function $\Tilde{h}$ satisfies \textbf{robust confidence-prediction calibration} on $\mathcal{D}^{\mathcal{T}}$ if for any confidence value $q \in [0, 1]$, any class $k \in [1, \cdots ,K]$, and the over-confident class $k'$, there exists a weighting constant $b \geq 1$ and corresponding $0 \leq a \leq 1$ that satisfies:
    \begin{equation}
        p({Y} = {k'} | \Tilde{h}_{k'} ({X}) = q) = a q,
    \end{equation}
    and
    \begin{equation}
        p({Y} = {k} | \Tilde{h}_{k} ({X}) = q) = b q \;\; \text{for $k \neq k'$}.
    \end{equation}
\end{definition}

Robust confidence-prediction calibration adjusts the over-confident expectation function $\Tilde{h}$ to have a lower probability on the misclassified class $k'$ via multiplying $a \leq 1$. Note that we can easily expand Definition~\ref{def:scpc} for multiple over-confident classes. Then, we estimate the test error with Theorem~\ref{thm:shifted-class-wise} (detailed proof in the Appendix~\ref{sec:app-shifted-class-wise}).

\begin{theorem}[Robust Disagreement Equality]\label{thm:shifted-class-wise}
    If the hypothesis space ${\mathcal{H}}_{\mathcal{A}}$ and corresponding expectation function $\Tilde{h}$ satisfies dropout independence and robust confidence-prediction calibration with a weighting constant $b$, prediction disagreement with dropouts (PDD) approximates the test error over ${\mathcal{H}}_{\mathcal{A}}$:
    \begin{equation}
        \mathbb{E}_{h \sim {\mathcal{H}}_{\mathcal{A}}} [{\tt{Err}}_{\mathcal{D}^{\mathcal{T}}}(h)] =
        b \; \mathbb{E}_{h \sim {\mathcal{H}}_{\mathcal{A}}} [{\tt{PDD}}_{\mathcal{D}^{\mathcal{T}}}(h)] - C,
    \end{equation}
    where
    \begin{equation}
        C = \int_{q \in [0, 1]} {(b-a)} \; q (1 - q) \; p (\Tilde{h}_{k'} ({X}) = q) dq.
    \end{equation}
\end{theorem}

\subsection{Accuracy Estimation for TTA}

With Theorem~\ref{thm:shifted-class-wise}, we propose an empirical approach to estimate the single model test error. 
Our experiments show that a single model's disagreement (and the test error) lies close to the robust disagreement equality. 
This aligns with the previous finding that a single pair of differently-trained models' disagreement rate (and the test error) lies close to the disagreement equality~\cite{disagreement}. Therefore, we approximate a single model test error as:
\begin{equation}\label{eq:general}
    {\tt{Err}}_{\mathcal{D}^{\mathcal{T}}}(h) \approx
    b \; {\tt{PDD}}_{\mathcal{D}^{\mathcal{T}}}(h),
\end{equation}
where we omit $C$ due to the insufficient information regarding the true value of $p (\Tilde{h}_{k'} ({X}) = q)$. Note that $C \approx 0$ for models with calibration.

Now, we discuss selecting a proper weighting constant $b$. Note that a desirable $b$ should dynamically suppress the over-confident expectation function depending on the context so that the confidence-prediction calibration assumption holds. To this end, we use the skewness of the predicted outputs as an indicator of model over-confidence. Out intuition is based on the observation that the predicted class distribution is highly skewed when the adaptation fails (Figure~\ref{fig:acc_conf_dist}), which aligns with the findings from prior studies~\cite{ood_energy, ood_baseline}. Specifically, we estimate the skewness of predictions by calculating the entropy ($\tt Ent$) of the batch-aggregated softmax values from the dropout inferences over a test batch $\mathbf{X}_t$:
\begin{equation}
\begin{aligned}
     E^{{\tt avg}} &= {\tt {Ent}} \left( \frac{1}{N} \sum_{i=1}^{N}  \frac{1}{|\mathbf{X}_t|} \sum_{\mathbf{x} \in \mathbf{X}_t} {f}(\mathbf{x}; \Theta^{{\tt dropout}_i}) \right),
\end{aligned}
\end{equation}
where $E^{{\tt avg}}$ would maximize as $E^{\tt{max}} = {\tt Ent} (\vec{1}_K / K)$ with uniform predictions among the batch (\eg, no failures); while the minimum value would be $0$ when entire batch predicts a single class (\eg, adaptation failures).

We then model $b$ with $E^{{\tt avg}}$ as:
\begin{equation}\label{eq:approx-b}
\begin{aligned}
    & b = \left( \frac{E^{{\tt avg}}}{E^{{\tt max}}} \right)^{-\alpha},
\end{aligned}
\end{equation}
where $\alpha \in [0, \infty)$ is a hyperparameter.
If the adaptation does not fail, predictions are uniformly distributed as $E^{{\tt avg}} = E^{{\tt max}}$ and $b = 1$. Note that $a=b=1$ drives Theorem~\ref{thm:shifted-class-wise} to be equivalent to Theorem~\ref{thm:class-wise}. We found that modeling $b$ with the average batch-wise entropy effectively corrects the correlation between confidence and prediction probability, as illustrated in Figure~\ref{fig:acc_corr} (blue dots).

Finally, with Equation~\ref{eq:general} and Equation~\ref{eq:approx-b}, we propose \textbf{Accuracy Estimation for TTA (\system{})}:
\begin{equation}\label{eq:wpdd}
    {\tt{Err}}_{\mathcal{D}^{\mathcal{T}}}(h) \approx
    \left( \frac{E^{{\tt avg}}}{E^{{\tt max}}} \right)^{-\alpha} {\tt{PDD}}_{\mathcal{D}^{\mathcal{T}}}(h).
\end{equation}

Observe that $\alpha=0$ and $\infty$ result in ${\tt{Err}}_{\mathcal{D}^{\mathcal{T}}}(h)={\tt{PDD}}_{\mathcal{D}^{\mathcal{T}}}(h)$ and ${\tt{Err}}_{\mathcal{D}^{\mathcal{T}}}(h)=1$, respectively. Setting a small $\alpha$ would result in a lesser penalty with adaptation failures. On the other hand, choosing a high $\alpha$ would undesirably penalize model improvement cases. Our experiment found that accuracy estimation is not too sensitive to $\alpha$ (Figure~\ref{fig:alpha}), and we chose $\alpha=3$ for the other experiments.

\begin{algorithm}
\caption{\system{}: batchwise TTA accuracy estimation}\label{alg:spdd}
\begin{algorithmic}
\Require Test batch $\mathbf{X}_t$, model $f$, number of dropout inferences $N$
\State ${\tt PDD} \leftarrow 0$
\State $\mathbf{Y}^{{\tt avg}} \leftarrow \vec{0}$
\State $\hat{\mathbf{Y}} \leftarrow f(\mathbf{X}_t; \Theta)$

\For{$i \in \{1, \cdots, N\}$}
    \State $\hat{\mathbf{Y}}^{\textit{d}} \leftarrow f(\mathbf{X}_t; \Theta^{{\tt dropout}_i})$
    \State $\mathbf{Y}^{{\tt avg}} \leftarrow \mathbf{Y}^{{\tt avg}} + {\tt Avg}({\hat{\mathbf{Y}}^{\textit{d}}\textit{\textbf{}}}) $ 
    \State ${\tt PDD} \leftarrow {\tt PDD} + {\tt Avg}({\mathbb{1} [ \argmax (\hat{\mathbf{Y}}) \neq \argmax  (\hat{\mathbf{Y}}^{\textit{d}})]})$
\EndFor

\State $\mathbf{Y}^{{\tt avg}} \leftarrow \frac{1}{N} \mathbf{Y}^{{\tt avg}}$
\State ${\tt PDD} \leftarrow \frac{1}{N}{\tt PDD}$ \Comment{Avg. over dropouts}

\State $E^{{\tt avg}} \leftarrow {\tt {Ent}} (\mathbf{Y}^{{\tt avg}})$ \Comment{Entropy of avg. batch}
\State ${\tt Err} \leftarrow \left( \frac{E^{{\tt avg}}}{E^{{\tt max}}} \right)^{-\alpha} {\tt PDD}$ \Comment{${\tt{Err}}_{\mathcal{D}^{\mathcal{T}}}(h)$}

\State ${\tt Acc} \leftarrow 1 - {\tt Err}$ 
\end{algorithmic}
\end{algorithm}

We summarize the accuracy estimation procedure in Algorithm~\ref{alg:spdd}. We first infer with the adapted model for the current test batch $\mathbf{X}_t$. Then, we repeatedly perform dropout inference sampling.
With $N$ samples from dropout inferences, we estimate the entropy of the batch-aggregated softmax output $E^{{\tt avg}}$. Finally, we calculated the expected error of the model by \system{}. We apply the exponential moving average to the final accuracy estimation for stable error estimation.

\section{Experiments}\label{sec:experiment}

\begin{table*}[t]
\centering
\caption{Mean absolute error (MAE) (\%) of the accuracy estimation on fully TTA (adapting to each corruption type). \textbf{Bold} numbers are the lowest error. Averaged over three different random seeds for 15 types of corruption.}
\vspace{-0.2cm}
\label{tab:main_experiment_iid}
\smaller
\begin{tabularx}{\textwidth}{ll*{6}Y>{\columncolor[HTML]{EDEEFF}}Y}
\Xhline{2\arrayrulewidth}
\addlinespace[0.08cm] 
 &  & \multicolumn{6}{c}{TTA Method} & \cellcolor[HTML]{FFFFFF} \\ \cline{3-8}
\addlinespace[0.08cm] 
\multirow{-2}{*}{Dataset} & \multirow{-2}{*}{Method} & TENT~\cite{tent} & EATA~\cite{eata} & SAR~\cite{sar} & CoTTA~\cite{cotta} & RoTTA~\cite{rotta} & SoTTA~\cite{sotta} & {\cellcolor[HTML]{EDEEFF}Avg. ($\downarrow$)} \\ \hline
 & SrcValid & 18.37 \scalebox{\std}{± 0.29} & 14.37 \scalebox{\std}{± 0.33} & 21.28 \scalebox{\std}{± 0.27} & 18.43 \scalebox{\std}{± 0.16} & 20.35 \scalebox{\std}{± 1.31} & 13.13 \scalebox{\std}{± 0.85} & 17.66 \scalebox{\std}{± 0.24} \\
 & SoftmaxScore~\cite{softmaxscore} & 6.26 \scalebox{\std}{± 0.49} & 4.78 \scalebox{\std}{± 0.12} & 5.21 \scalebox{\std}{± 0.22} & 10.96 \scalebox{\std}{± 0.28} & 6.01 \scalebox{\std}{± 0.23} & 4.97 \scalebox{\std}{± 0.50} & 6.37 \scalebox{\std}{± 0.10} \\
 & GDE~\cite{disagreement} & 18.69 \scalebox{\std}{± 0.28} & 16.95 \scalebox{\std}{± 0.22} & 21.25 \scalebox{\std}{± 0.27} & 14.50 \scalebox{\std}{± 0.03} & 23.27 \scalebox{\std}{± 0.43} & 16.45 \scalebox{\std}{± 0.21} & 18.52 \scalebox{\std}{± 0.13} \\
 & AdvPerturb~\cite{advperturb} & 23.06 \scalebox{\std}{± 1.17} & 24.97 \scalebox{\std}{± 1.00} & 21.89 \scalebox{\std}{± 0.95} & 18.00 \scalebox{\std}{± 0.82} & 19.35 \scalebox{\std}{± 0.99} & 23.68 \scalebox{\std}{± 0.85} & 21.83 \scalebox{\std}{± 0.92} \\
\multirow{-5}{*}{\begin{tabular}[c]{@{}l@{}}Fully\\ CIFAR10-C\end{tabular}} & \cellcolor[HTML]{E2FFE1}\system{} & \cellcolor[HTML]{E2FFE1}4.00 \scalebox{\std}{± 0.03} & \cellcolor[HTML]{E2FFE1}3.87 \scalebox{\std}{± 0.14} & \cellcolor[HTML]{E2FFE1}3.89 \scalebox{\std}{± 0.07} & \cellcolor[HTML]{E2FFE1}6.83 \scalebox{\std}{± 0.47} & \cellcolor[HTML]{E2FFE1}6.44 \scalebox{\std}{± 1.35} & \cellcolor[HTML]{E2FFE1}5.28 \scalebox{\std}{± 0.87} & \cellcolor[HTML]{E2FFE1}\textbf{5.05 \scalebox{\std}{± 0.46}} \\ \hline
 & SrcValid & 38.96 \scalebox{\std}{± 0.22} & 10.71 \scalebox{\std}{± 0.31} & 42.68 \scalebox{\std}{± 0.21} & 44.58 \scalebox{\std}{± 0.30} & 23.50 \scalebox{\std}{± 0.51} & 19.34 \scalebox{\std}{± 0.63} & 29.96 \scalebox{\std}{± 0.09} \\
 & SoftmaxScore~\cite{softmaxscore} & 17.34 \scalebox{\std}{± 0.10} & 27.86 \scalebox{\std}{± 1.11} & 24.56 \scalebox{\std}{± 0.25} & 34.50 \scalebox{\std}{± 0.35} & 24.18 \scalebox{\std}{± 0.19} & 23.98 \scalebox{\std}{± 0.21} & 25.40 \scalebox{\std}{± 0.23} \\
 & GDE~\cite{disagreement} & 40.11 \scalebox{\std}{± 0.05} & 71.53 \scalebox{\std}{± 2.12} & 42.51 \scalebox{\std}{± 0.23} & 33.21 \scalebox{\std}{± 0.24} & 48.02 \scalebox{\std}{± 0.56} & 34.24 \scalebox{\std}{± 0.12} & 44.94 \scalebox{\std}{± 0.23} \\
 & AdvPerturb~\cite{advperturb} & 24.17 \scalebox{\std}{± 0.41} & 8.22 \scalebox{\std}{± 0.56} & 22.91 \scalebox{\std}{± 0.60} & 20.53 \scalebox{\std}{± 0.14} & 17.84 \scalebox{\std}{± 0.65} & 25.77 \scalebox{\std}{± 0.47} & 19.91 \scalebox{\std}{± 0.26} \\
\multirow{-5}{*}{\begin{tabular}[c]{@{}l@{}}Fully\\ CIFAR100-C\end{tabular}} & \cellcolor[HTML]{E2FFE1}\system{} & \cellcolor[HTML]{E2FFE1}6.89 \scalebox{\std}{± 0.15} & \cellcolor[HTML]{E2FFE1}20.15 \scalebox{\std}{± 1.70} & \cellcolor[HTML]{E2FFE1}6.54 \scalebox{\std}{± 0.15} & \cellcolor[HTML]{E2FFE1}6.05 \scalebox{\std}{± 0.12} & \cellcolor[HTML]{E2FFE1}6.88 \scalebox{\std}{± 0.10} & \cellcolor[HTML]{E2FFE1}5.29 \scalebox{\std}{± 0.18} & \cellcolor[HTML]{E2FFE1}\textbf{8.63 \scalebox{\std}{± 0.24}} \\ \hline
 & SrcValid & 39.13 \scalebox{\std}{± 0.89} & 35.89 \scalebox{\std}{± 0.79} & 29.77 \scalebox{\std}{± 0.94} & 41.09 \scalebox{\std}{± 0.53} & 10.28 \scalebox{\std}{± 0.28} & 16.00 \scalebox{\std}{± 0.33} & 28.69 \scalebox{\std}{± 0.54} \\
 & SoftmaxScore~\cite{softmaxscore} & 20.67 \scalebox{\std}{± 0.01} & 21.06 \scalebox{\std}{± 0.03} & 24.42 \scalebox{\std}{± 0.08} & 19.62 \scalebox{\std}{± 0.02} & 21.03 \scalebox{\std}{± 0.04} & 23.60 \scalebox{\std}{± 0.07} & 21.73 \scalebox{\std}{± 0.03} \\
 & GDE~\cite{disagreement} & 70.58 \scalebox{\std}{± 0.01} & 66.17 \scalebox{\std}{± 0.07} & 63.48 \scalebox{\std}{± 0.03} & 72.76 \scalebox{\std}{± 0.02} & 66.39 \scalebox{\std}{± 0.04} & 52.74 \scalebox{\std}{± 0.02} & 65.35 \scalebox{\std}{± 0.02} \\
 & AdvPerturb~\cite{advperturb} & 12.56 \scalebox{\std}{± 0.03} & 14.52 \scalebox{\std}{± 0.01} & 18.76 \scalebox{\std}{± 0.06} & 11.05 \scalebox{\std}{± 0.02} & 12.93 \scalebox{\std}{± 0.04} & 22.90 \scalebox{\std}{± 0.02} & 15.45 \scalebox{\std}{± 0.02} \\
\multirow{-5}{*}{\begin{tabular}[c]{@{}l@{}}Fully\\ ImageNet-C\end{tabular}} & \cellcolor[HTML]{E2FFE1}\system{} & \cellcolor[HTML]{E2FFE1}6.14 \scalebox{\std}{± 0.03} & \cellcolor[HTML]{E2FFE1}6.48 \scalebox{\std}{± 0.02} & \cellcolor[HTML]{E2FFE1}6.43 \scalebox{\std}{± 0.09} & \cellcolor[HTML]{E2FFE1}6.02 \scalebox{\std}{± 0.03} & \cellcolor[HTML]{E2FFE1}14.82 \scalebox{\std}{± 0.01} & \cellcolor[HTML]{E2FFE1}17.40 \scalebox{\std}{± 0.26} & \cellcolor[HTML]{E2FFE1}\textbf{9.55 \scalebox{\std}{± 0.07}} \\ 
\Xhline{2\arrayrulewidth}
\end{tabularx}
\end{table*}

\begin{table*}[t]
\centering
\caption{Mean absolute error (MAE) (\%) of the accuracy estimation on continual TTA (continuously adapting to 15 consecutive corruptions). \textbf{Bold} numbers are the lowest error. Averaged over three different random seeds for 15 types of corruption.}
\vspace{-0.2cm}
\label{tab:main_experiment_cont}
\smaller
\begin{tabularx}{\textwidth}{ll*{6}Y>{\columncolor[HTML]{EDEEFF}}Y}
\Xhline{2\arrayrulewidth}
\addlinespace[0.08cm] 
 &  & \multicolumn{6}{c}{TTA Method} & \cellcolor[HTML]{FFFFFF} \\ \cline{3-8}
\addlinespace[0.08cm] 
\multirow{-2}{*}{Dataset} & \multirow{-2}{*}{Method} & TENT~\cite{tent} & EATA~\cite{eata} & SAR~\cite{sar} & CoTTA~\cite{cotta} & RoTTA~\cite{rotta} & SoTTA~\cite{sotta} & {\cellcolor[HTML]{EDEEFF}Avg. ($\downarrow$)} \\ \hline
 & SrcValid & 10.84 \scalebox{\std}{± 1.83} & 11.06 \scalebox{\std}{± 0.11} & 21.29 \scalebox{\std}{± 0.26} & 18.30 \scalebox{\std}{± 0.25} & 13.37 \scalebox{\std}{± 0.89} & 9.40 \scalebox{\std}{± 0.85} & 14.04 \scalebox{\std}{± 0.58} \\
 & SoftmaxScore~\cite{softmaxscore} & 41.10 \scalebox{\std}{± 11.66} & 15.40 \scalebox{\std}{± 4.73} & 5.21 \scalebox{\std}{± 0.22} & 12.96 \scalebox{\std}{± 0.37} & 12.57 \scalebox{\std}{± 0.43} & 4.37 \scalebox{\std}{± 0.09} & 15.27 \scalebox{\std}{± 2.51} \\
 & GDE~\cite{disagreement} & 46.29 \scalebox{\std}{± 10.93} & 26.44 \scalebox{\std}{± 5.16} & 21.25 \scalebox{\std}{± 0.27} & 14.69 \scalebox{\std}{± 0.15} & 17.50 \scalebox{\std}{± 0.30} & 17.03 \scalebox{\std}{± 0.70} & 23.87 \scalebox{\std}{± 2.43} \\
 & AdvPerturb~\cite{advperturb} & 15.56 \scalebox{\std}{± 1.53} & 20.93 \scalebox{\std}{± 2.83} & 21.88 \scalebox{\std}{± 0.93} & 17.79 \scalebox{\std}{± 0.74} & 22.95 \scalebox{\std}{± 0.82} & 23.63 \scalebox{\std}{± 0.78} & 20.45 \scalebox{\std}{± 1.17} \\
\multirow{-5}{*}{\begin{tabular}[c]{@{}l@{}}Continual\\ CIFAR10-C\end{tabular}} & \cellcolor[HTML]{E2FFE1}\system{} & \cellcolor[HTML]{E2FFE1}9.05 \scalebox{\std}{± 1.02} & \cellcolor[HTML]{E2FFE1}7.13 \scalebox{\std}{± 3.33} & \cellcolor[HTML]{E2FFE1}3.89 \scalebox{\std}{± 0.06} & \cellcolor[HTML]{E2FFE1}5.82 \scalebox{\std}{± 0.30} & \cellcolor[HTML]{E2FFE1}5.36 \scalebox{\std}{± 1.22} & \cellcolor[HTML]{E2FFE1}4.73 \scalebox{\std}{± 0.34} & \cellcolor[HTML]{E2FFE1}\textbf{6.00 \scalebox{\std}{± 0.35}} \\ \hline
 & SrcValid & 11.00 \scalebox{\std}{± 0.58} & 1.68 \scalebox{\std}{± 0.18} & 38.20 \scalebox{\std}{± 0.22} & 46.09 \scalebox{\std}{± 0.38} & 19.43 \scalebox{\std}{± 1.17} & 17.16 \scalebox{\std}{± 1.57} & 22.32 \scalebox{\std}{± 0.52} \\
 & SoftmaxScore~\cite{softmaxscore} & 58.29 \scalebox{\std}{± 1.82} & 76.58 \scalebox{\std}{± 0.71} & 24.05 \scalebox{\std}{± 0.29} & 36.27 \scalebox{\std}{± 0.68} & 27.19 \scalebox{\std}{± 0.12} & 21.89 \scalebox{\std}{± 0.35} & 40.71 \scalebox{\std}{± 0.43} \\
 & GDE~\cite{disagreement} & 80.87 \scalebox{\std}{± 1.29} & 94.01 \scalebox{\std}{± 0.43} & 39.21 \scalebox{\std}{± 0.22} & 35.43 \scalebox{\std}{± 0.30} & 41.68 \scalebox{\std}{± 0.45} & 35.29 \scalebox{\std}{± 0.27} & 54.41 \scalebox{\std}{± 0.18} \\
 & AdvPerturb~\cite{advperturb} & 10.12 \scalebox{\std}{± 0.24} & 1.97 \scalebox{\std}{± 0.33} & 24.93 \scalebox{\std}{± 0.57} & 19.62 \scalebox{\std}{± 0.15} & 21.18 \scalebox{\std}{± 0.71} & 25.12 \scalebox{\std}{± 0.39} & 17.16 \scalebox{\std}{± 0.32} \\
\multirow{-5}{*}{\begin{tabular}[c]{@{}l@{}}Continual\\ CIFAR100-C\end{tabular}} & \cellcolor[HTML]{E2FFE1}\system{} & \cellcolor[HTML]{E2FFE1}5.85 \scalebox{\std}{± 0.36} & \cellcolor[HTML]{E2FFE1}4.18 \scalebox{\std}{± 0.82} & \cellcolor[HTML]{E2FFE1}6.67 \scalebox{\std}{± 0.12} & \cellcolor[HTML]{E2FFE1}6.55 \scalebox{\std}{± 0.17} & \cellcolor[HTML]{E2FFE1}5.86 \scalebox{\std}{± 0.10} & \cellcolor[HTML]{E2FFE1}5.32 \scalebox{\std}{± 0.18} & \cellcolor[HTML]{E2FFE1}\textbf{5.74 \scalebox{\std}{± 0.13}} \\ \hline
 & SrcValid & 33.30 \scalebox{\std}{± 0.93} & 36.42 \scalebox{\std}{± 0.76} & 22.30 \scalebox{\std}{± 0.55} & 41.06 \scalebox{\std}{± 0.54} & 9.56 \scalebox{\std}{± 0.26} & 14.28 \scalebox{\std}{± 0.28} & 26.15 \scalebox{\std}{± 0.53} \\
 & SoftmaxScore~\cite{softmaxscore} & 19.34 \scalebox{\std}{± 0.02} & 20.16 \scalebox{\std}{± 0.05} & 21.91 \scalebox{\std}{± 0.16} & 19.63 \scalebox{\std}{± 0.01} & 17.56 \scalebox{\std}{± 0.08} & 19.67 \scalebox{\std}{± 0.50} & 19.71 \scalebox{\std}{± 0.53} \\
 & GDE~\cite{disagreement} & 68.30 \scalebox{\std}{± 0.01} & 66.58 \scalebox{\std}{± 0.03} & 64.36 \scalebox{\std}{± 0.15} & 72.81 \scalebox{\std}{± 0.07} & 73.76 \scalebox{\std}{± 0.22} & 55.76 \scalebox{\std}{± 0.45} & 66.93 \scalebox{\std}{± 0.14} \\
 & AdvPerturb~\cite{advperturb} & 14.82 \scalebox{\std}{± 0.02} & 14.15 \scalebox{\std}{± 0.06} & 19.17 \scalebox{\std}{± 0.14} & 11.06 \scalebox{\std}{± 0.02} & 11.05 \scalebox{\std}{± 0.05} & 20.83 \scalebox{\std}{± 0.39} & 15.18 \scalebox{\std}{± 0.09} \\
\multirow{-5}{*}{\begin{tabular}[c]{@{}l@{}}Continual\\ ImageNet-C\end{tabular}} & \cellcolor[HTML]{E2FFE1}\system{} & \cellcolor[HTML]{E2FFE1}5.66 \scalebox{\std}{± 0.05} & \cellcolor[HTML]{E2FFE1}6.73 \scalebox{\std}{± 0.03} & \cellcolor[HTML]{E2FFE1}6.68 \scalebox{\std}{± 0.04} & \cellcolor[HTML]{E2FFE1}5.98 \scalebox{\std}{± 0.04} & \cellcolor[HTML]{E2FFE1}11.19 \scalebox{\std}{± 0.12} & \cellcolor[HTML]{E2FFE1}19.22 \scalebox{\std}{± 0.79} & \cellcolor[HTML]{E2FFE1}\textbf{9.24 \scalebox{\std}{± 0.14}}
\\
\Xhline{2\arrayrulewidth}
\end{tabularx}
\end{table*}

We describe our experimental setup and present the results. Please refer to the Appendix~\ref{sec:app-exp-details} for further details.

\noindent\textbf{Scenario.} We consider both fully (non-continual) and continual test-time adaptation scenarios. In the fully TTA setting, target domains are each corruption type~\cite{tent}, while in the continual setting, the target domain continually changes to 15 different corruptions~\cite{cotta}. During adaptation, we calculate the accuracy estimation for every batch and report the mean absolute error between the ground-truth batch-wise accuracy. We ran experiments with three random seeds (0, 1, 2) and reported the average values. We use the test batch size 64 for all TTA baselines, with a memory size 64 for RoTTA~\cite{rotta} and SoTTA~\cite{sotta}. We specify further details of the hyperparameters in the Appendix~\ref{sec:app-tta-details}.

\begin{figure*}[t]
  \centering
    \begin{subfigure}[t]{1\linewidth}
    \centering
    \includegraphics[width=0.24\linewidth]{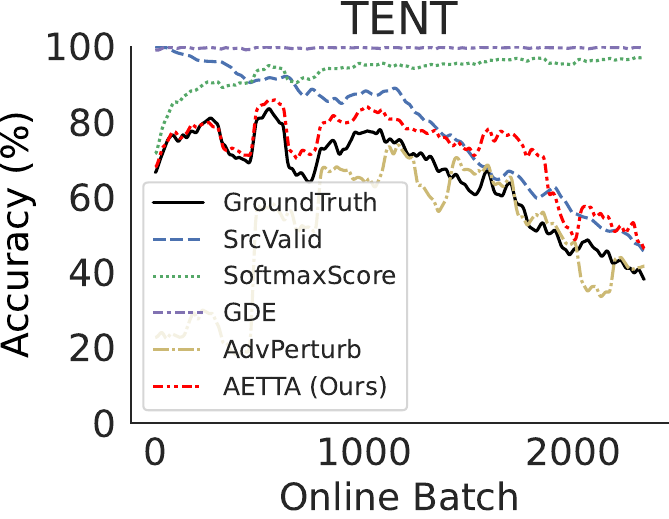}
    \includegraphics[width=0.24\linewidth]{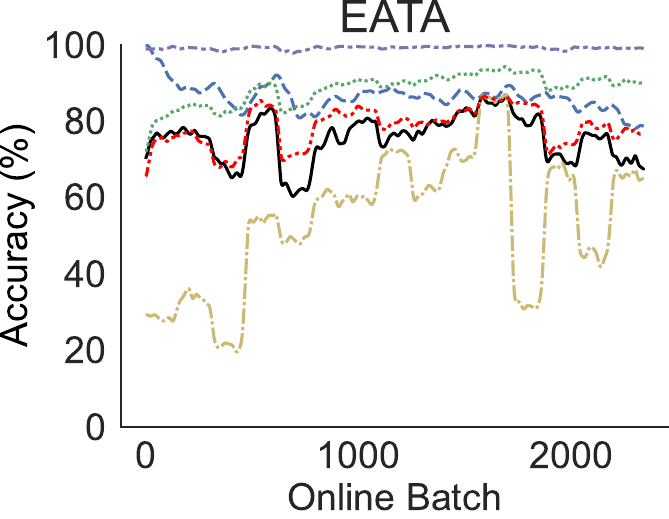}
    \includegraphics[width=0.24\linewidth]{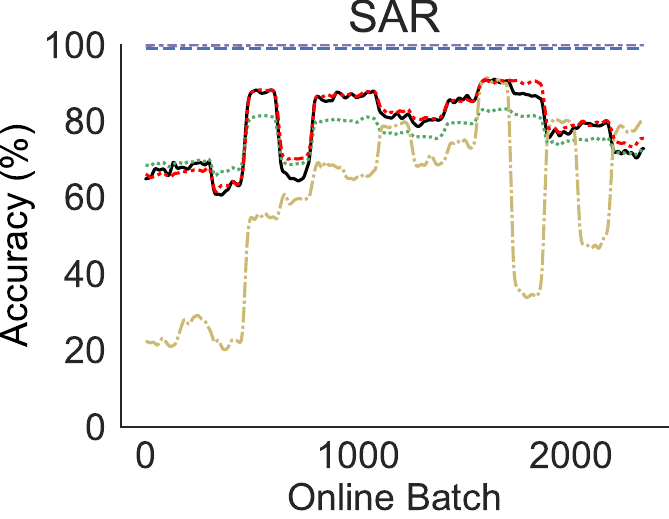}
    \includegraphics[width=0.24\linewidth]{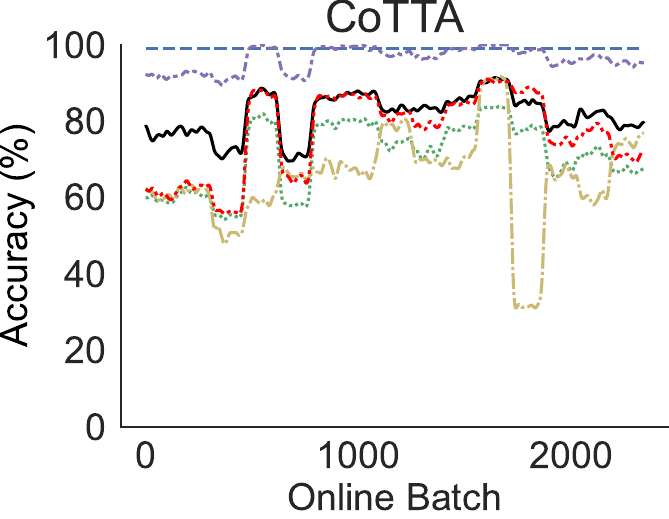}
    \caption{CIFAR10-C.}
    \end{subfigure}
    \begin{subfigure}[t]{1\linewidth}
    \centering
    \includegraphics[width=0.24\linewidth]{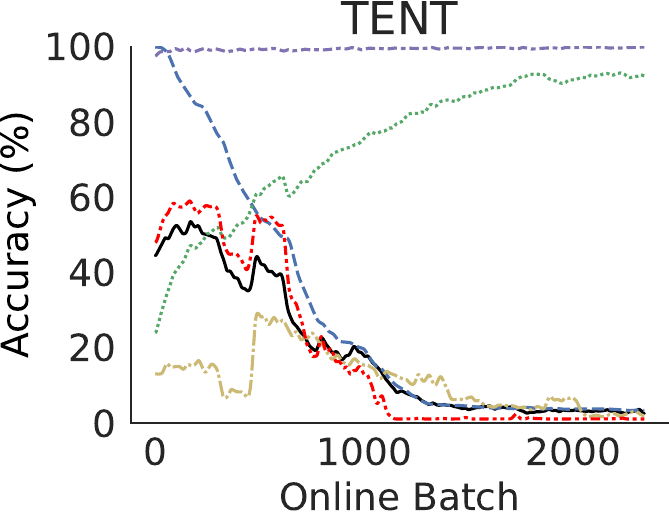}
    \includegraphics[width=0.24\linewidth]{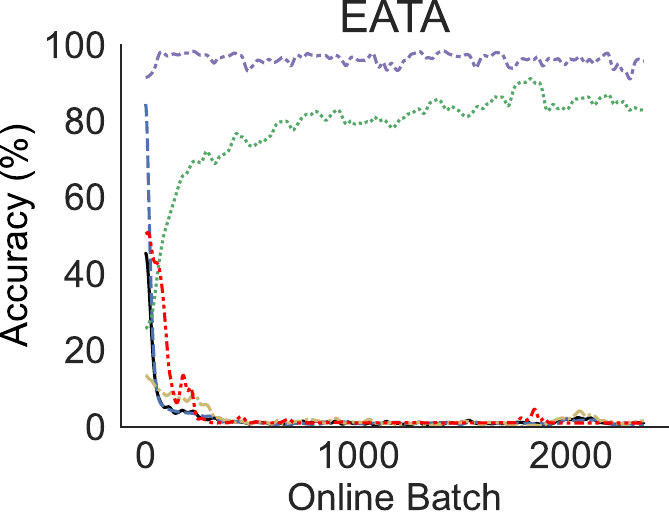}
    \includegraphics[width=0.24\linewidth]{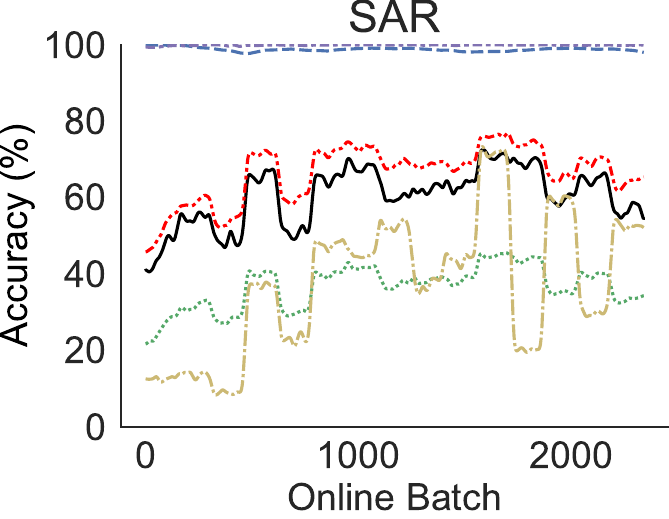}
    \includegraphics[width=0.24\linewidth]{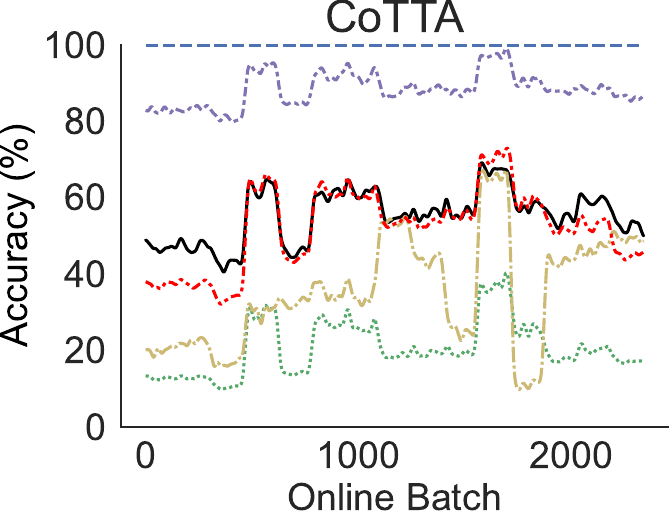}
    \caption{CIFAR100-C.}
    \end{subfigure}
    \begin{subfigure}[t]{1\linewidth}
    \centering
    \includegraphics[width=0.24\linewidth]{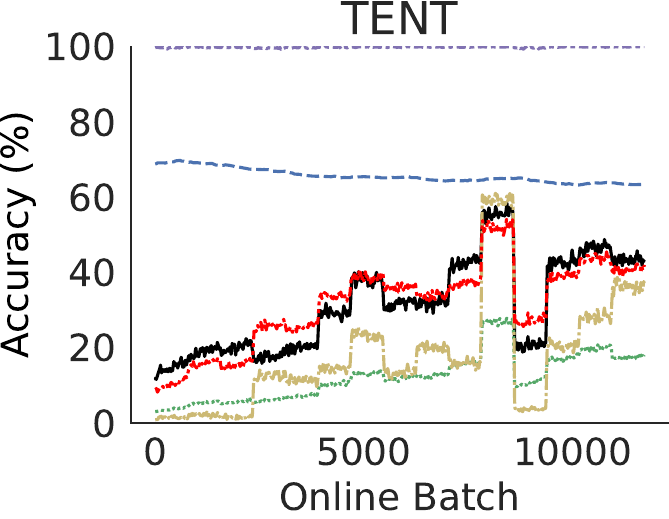}
    \includegraphics[width=0.24\linewidth]{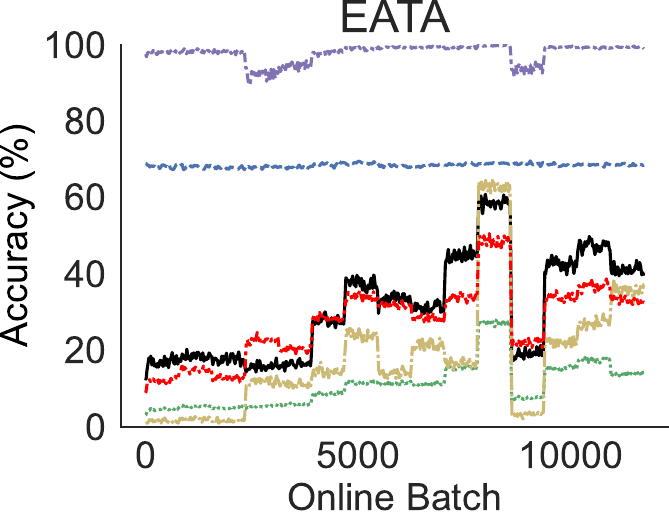}
    \includegraphics[width=0.24\linewidth]{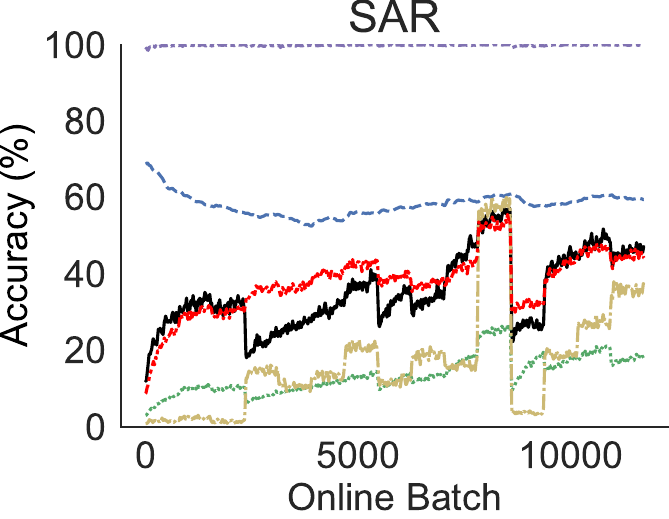}
    \includegraphics[width=0.24\linewidth]{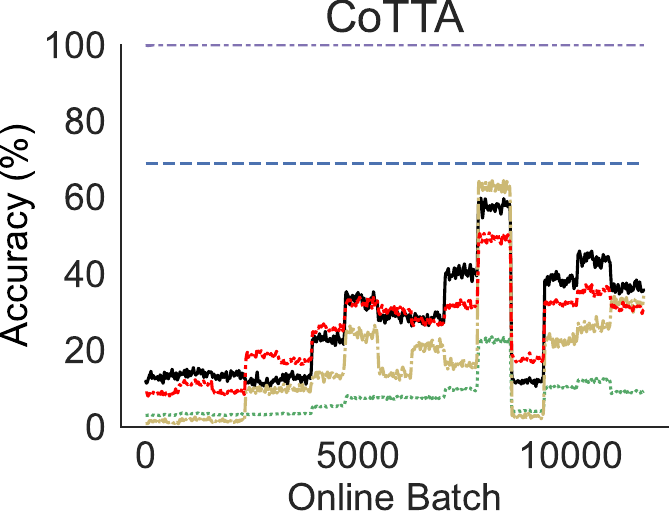}
    \caption{ImageNet-C.}
    \end{subfigure}
  \caption{Qualitative results on continual CIFAR10-C, CIFAR100-C, and ImageNet-C. 
  }\label{fig:qualitative}
\end{figure*}

\noindent\textbf{Datasets.} We use three standard benchmarks for test-time adaptation: \textbf{CIFAR10-C}, \textbf{CIFAR100-C}, and \textbf{ImageNet-C}~\cite{cifarc}. Each dataset contains 15 different corruptions with five levels of corruption, where we use corruption level 5. CIFAR10-C/CIFAR100-C/ImageNet-C contains 10/100/1,000 classes with 10,000/10,000/50,000 test data, respectively. We use pre-trained ResNet18~\cite{7780459} as an adaptation target, following a recent study~\cite{sotta}.

\noindent\textbf{TTA Methods.} We consider six state-of-the-art TTA methods. TENT~\cite{tent} updates BN parameters with entropy minimization. EATA~\cite{eata} utilizes entropy thresholding-based sample filtering and anti-forgetting regularization. SAR~\cite{sar} also adapts sample filtering with sharpness minimization~\cite{sam}. CoTTA~\cite{cotta} addresses the continual setting by augmentations and stochastic restoration of model weights to avoid catastrophic forgetting. RoTTA~\cite{rotta} adapts with robust batch normalization and category-balanced sampling with timeliness and uncertainty. SoTTA~\cite{sotta} utilizes high-confidence uniform-sampling and entropy-sharpness minimization for robust adaptation in noisy data streams~\cite{sam}.

\noindent\textbf{Accuracy Estimation Baselines.} 
We evaluate four distinct accuracy estimation baselines that could be applied to TTA settings:
SrcValid, SoftmaxScore, GDE, and AdvPerturb.
\begin{itemize}
    \item \textbf{SrcValid} is a widely used technique that validates performance by leveraging labeled source data. It computes the accuracy using a hold-out labeled source dataset to estimate the target performance. Importantly, the hold-out source data for validation were not used for training in other baselines to ensure they do not affect the model performance. Note that TTA usually assumes that source data are unavailable during test time; hence, this baseline is unrealistic in TTA. We nonetheless include SrcValid as one of our baselines to understand its performance when the source data are accessible.
    \item \textbf{SoftmaxScore}~\cite{softmaxscore} utilizes the confidence scores derived from the last softmax layer as the model's accuracy, which is also a widely used baseline~\cite{chen2021ensemble, chuang2020estimating}. It estimates the target domain accuracy by averaging softmax confidence scores computed from the current test batch. In addition, we apply temperature scaling~\cite{guo2017calibration} to improve the estimation performance~\cite{atc}.
    \item Generalization disagreement equality (\textbf{GDE})~\cite{disagreement} aims to estimate test accuracy by quantifying the (dis)agreement rate between predictions on a test batch generated by a pair of models. Since training multiple models is impractical, we compare the current adapted model and the previous model right before the adaptation. We also report a comparison with the original GDE and multiple pre-trained models in Appendix~\ref{sec:app-add-exp}.
    \item Adversarial perturbation (\textbf{AdvPerturb})~\cite{advperturb} also aims to estimate the OOD accuracy by calculating the agreement between the domain-adapted model and the source model, where adversarial perturbations on a test batch are applied to penalize the unconfident samples near the decision boundary. We note that the original paper aims to predict the accuracy of the source model, while our goal is to predict the accuracy of the adapted model.
\end{itemize}

\noindent\textbf{Results.} 
Table~\ref{tab:main_experiment_iid} and Table~\ref{tab:main_experiment_cont} show the results on the fully and continual TTA settings. We observe that none of the baselines could reliably predict the accuracy among different scenarios. On the other hand, \system{} achieves the lowest mean absolute error, including adaptation failure cases (\eg, TENT in continual CIFAR10/100-C). On average, \system{} outperforms baselines by 19.8\%p, validating the effectiveness of our robust prediction disagreement in diverse scenarios. More details are in the Appendix~\ref{sec:app-result-details}.

\noindent\textbf{Qualitative Analysis.}
We qualitatively analyze the results of the baselines and \system{} to understand the behavior. Figure~\ref{fig:qualitative} visualizes the ground-truth accuracy and the estimated accuracy from the baselines and \system{} under adaptation failure and non-failure cases. The Gaussian filter is applied for visualization. 
We observe that \system{} generally shows a reliable estimation of the ground-truth accuracy in diverse scenarios (fully and continual) and datasets (CIFAR10/100-C and ImageNet-C).
SrcValid correctly estimated when model accuracy decreases;
however, it consistently predicted high accuracy when the adaptation did not fail. This limitation might be due to the distributional gap between source and target data.
SoftmaxScore~\cite{softmaxscore} captures the trend of ground-truth accuracy in some cases, 
but it overestimates the accuracy 
when the model accuracy drops. This is mostly due to the over-confident predictions from the model. 
GDE~\cite{disagreement} showed to constantly predict high values among different TTA methods. 
Note that GDE was originally designed to utilize various pre-trained models. To use GDE in TTA, we utilize adapted models sampled at different stages of adaptation. The result suggests that utilizing multiple models from the single stochastic learning process might not be sufficient to consist of independent and identically distributed (i.i.d.) ensembles, leading to inaccurate estimation. 
AdvPerturb~\cite{advperturb} shows accuracy estimations when ground-truth accuracy decreases but shows high errors in other cases. We believe this happens because it
aims to evaluate the performance of the source model, not the adapted model. We found similar patterns were observed with different TTA methods.\\

\begin{figure}[t]
    \centering
    \begin{subfigure}[t]{0.495\linewidth}
    \centering
    \includegraphics[width=0.95\linewidth]{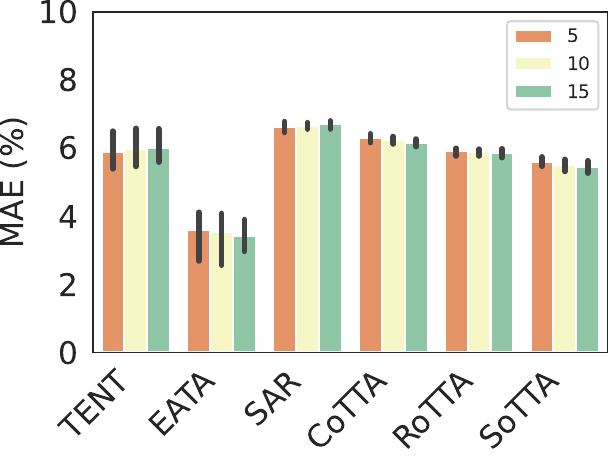}
    \caption{Number of dropout inferences $N$.}\label{fig:n}
    \end{subfigure}
    \begin{subfigure}[t]{0.495\linewidth}
    \centering
    \includegraphics[width=0.95\linewidth]{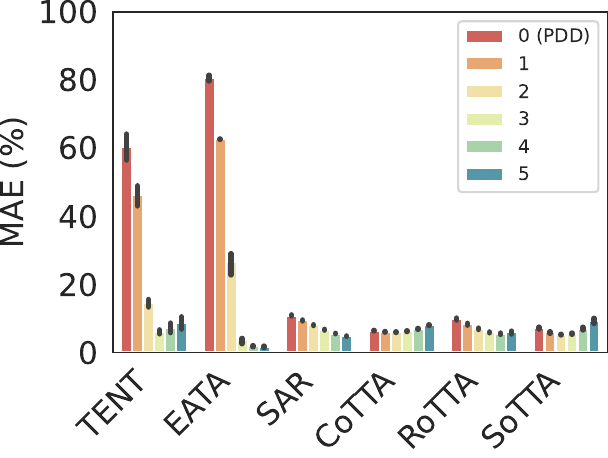}
    \caption{Scaling hyperparameter $\alpha$.}\label{fig:alpha}
    \end{subfigure}
    \caption{Impact of hyperparameters on the accuracy estimation performance.}
\end{figure}

\noindent\textbf{Impact of Hyperparameter $N$.}
The number of dropout inferences, $N$, is a hyperparameter for calculating the test error. We conducted an ablation study in continual CIFAR100-C with varying $N \in \{5, 10, 15\}$. As shown in Figure~\ref{fig:n}, we found the effect of hyperparameter $N$ is negligible. We interpret this result as the effect of calculating prediction disagreement over sufficient batch size with dropout independence, which could reduce the probabilistic variances from dropout inference sampling.
We adopt a single value of $N=10$ for the other experiments.

\noindent\textbf{Impact of Hyperparameter $\alpha$.}
We investigate the impact of $\alpha$, a hyperparameter to control the strength of robust confidence-prediction calibration. We conduct an ablation study in continual CIFAR100-C with varying $\alpha \in \{0, \dots, 5\}$, where $\alpha = 0$ indicates no weighting, thus $\tt Err = PDD$. Figure~\ref{fig:alpha} shows the result. Note that estimations are often inaccurate when $\alpha = 0$, which shows the importance of our robust equality. Setting a reasonable $\alpha$ is important to predict failed adaptation cases (TENT and EATA) properly, but it is generally robust after certain values. We adopt $\alpha = 3$ for the other experiments.

\section{Case Study: Model Recovery} \label{sec:recovery}

\begin{table*}[t]
\centering
\caption{Average accuracy improvement (\%p) with model recovery. \textbf{Bold} number is the highest improvement. Averaged over three different random seeds for 15 types of corruption.}
\vspace{-0.2cm}
\label{tab:app_reset}
\smaller
\begin{tabularx}{\linewidth}{l*{7}Y}
\Xhline{2\arrayrulewidth}
\addlinespace[0.08cm] 
 & \multicolumn{6}{c}{TTA Method} & \\ \cline{2-7}
\addlinespace[0.08cm] 
\multirow{-2}{*}{Method} & TENT~\cite{tent} & EATA~\cite{eata} & SAR~\cite{sar} & CoTTA~\cite{cotta} & RoTTA~\cite{rotta} & SoTTA~\cite{sotta} & {\cellcolor[HTML]{EDEEFF}Avg. ($\uparrow$)} \\ \hline
Episodic~\cite{memo} & 33.58 \scalebox{\std}{± 1.04} & 51.28 \scalebox{\std}{± 0.52} & -7.00 \scalebox{\std}{± 0.26} & 1.65 \scalebox{\std}{± 0.10} & -22.57 \scalebox{\std}{± 0.85} & -26.40 \scalebox{\std}{± 0.51} & \cellcolor[HTML]{EDEEFF}5.09 \scalebox{\std}{± 0.24} \\
MRS~\cite{sar} & 24.12 \scalebox{\std}{± 2.11} & 0.00 \scalebox{\std}{± 0.00} & 0.00 \scalebox{\std}{± 0.00} & 0.00 \scalebox{\std}{± 0.00} & -1.97 \scalebox{\std}{± 2.23} & 0.00 \scalebox{\std}{± 0.00} & \cellcolor[HTML]{EDEEFF}3.69 \scalebox{\std}{± 0.22}\\
Stochastic~\cite{cotta} & 35.93 \scalebox{\std}{± 0.78} & -0.01 \scalebox{\std}{± 0.47} & -2.00 \scalebox{\std}{± 0.48} & 0.00 \scalebox{\std}{± 0.00} & -2.55 \scalebox{\std}{± 0.49} & 0.35 \scalebox{\std}{± 0.51} & \cellcolor[HTML]{EDEEFF}5.29 \scalebox{\std}{± 0.19}\\
FisherStochastic~\cite{petal} & 40.27 \scalebox{\std}{± 1.29} & 0.12 \scalebox{\std}{± 1.16} & -4.85 \scalebox{\std}{± 0.13} & 0.13 \scalebox{\std}{± 0.03} & -2.89 \scalebox{\std}{± 0.13} & -1.36 \scalebox{\std}{± 0.51} & \cellcolor[HTML]{EDEEFF}5.24 \scalebox{\std}{± 0.29}\\
DistShift & 38.93 \scalebox{\std}{± 1.15} & 22.17 \scalebox{\std}{± 2.38} & -3.25 \scalebox{\std}{± 0.10} & 1.51 \scalebox{\std}{± 0.09} & -7.63 \scalebox{\std}{± 0.23} & 0.68 \scalebox{\std}{± 0.19} & \cellcolor[HTML]{EDEEFF}8.74 \scalebox{\std}{± 0.55} \\
\rowcolor[HTML]{E2FFE1} 
\cellcolor[HTML]{E2FFE1}\system{}& 36.79 \scalebox{\std}{± 1.20} & 48.64 \scalebox{\std}{± 0.74} & -5.66 \scalebox{\std}{± 0.20} & 1.64 \scalebox{\std}{± 0.11} & -6.03 \scalebox{\std}{± 0.89} & -4.97 \scalebox{\std}{± 1.58} & \textbf{11.73 \scalebox{\std}{± 0.34}} \\ 
\Xhline{2\arrayrulewidth}
\end{tabularx}
\end{table*}

The deployment of TTA algorithms encounters a significant challenge when exposed to extreme test streams, such as continuously changing corruptions~\cite{cotta}. Several TTA algorithms (\eg, TENT~\cite{tent}) were not designed to exhibit robustness under such extreme conditions. Consequently, the model weights are poorly updated, leading to performance degradation, even worse than the source model. Although recent studies attempt to manage dynamic test streams~\cite{note, cotta, sotta}, TTA algorithms are still susceptible to adaptation failures~\cite{rdumb}. To tackle the issue, we perform a case study of model recovery based on the accuracy estimation.\\

\noindent\textbf{Recovery Algorithm.} We introduce a simple reset algorithm based on our accuracy estimation with \system{}.
Our reset algorithm detects two cases: (1) consecutive low accuracies and (2) sudden accuracy drop. First, we reset the model if the five recent consecutive estimated accuracies (\eg, $t-4, \cdots, t$) are lower than the five previous consecutive estimations (\eg, $t-9, \cdots, t-5$). 
This way, we can detect the gradual degradation of TTA accuracy. Second, we apply hard lower-bound thresholding, which resets the model if the estimated accuracy is below the threshold (\eg, 0.2). This could prevent catastrophic failure of TTA algorithms.\\

\noindent\textbf{Baselines.} Some TTA studies covered the model recovery/reset as a part of the TTA algorithm: Episodic resetting (\textbf{Episodic})~\cite{memo}, where the model resets after every batch; Model Recovery Scheme (\textbf{MRS})~\cite{sar}, where the model resets when the moving average of entropy loss falls below a certain threshold; Stochastic restoration (\textbf{Stochastic})~\cite{cotta}, where a small number of model weights are stochastically restored to the initial weight of the source model; and Fisher information based restoration (\textbf{FisherStochastic})~\cite{petal}, which applies stochastic restoration for layer importance measured by Fisher information matrix. We also include a baseline (\textbf{DistShift}), which assumes that the model knows when the distribution changes and thus acts as an oracle. DistShift resets the model when the test data distribution (corruption) changes, which is not feasible in practice.

\noindent\textbf{Results.} Our simple recovery algorithm outperforms the baselines, including DistShift, which relies on an impractical assumption of knowing when the corruption changes. Episodic~\cite{memo} showed high accuracy improvements under adaptation failures; however, it prevents
continuous adaptation, even without adaptation failures. MRS~\cite{sar} fails to recover among various TTA methods due to the hard-coded threshold of loss value. Stochastic~\cite{cotta} and FisherStochastic~\cite{petal} show marginal improvements while failing to recover EATA. Our proposed reset algorithm successfully recovers from adaptation failures while minimizing the negative effect on TTA without failures. \\

\begin{figure}[t]
  \centering
    \begin{subfigure}[t]{1\linewidth}
    \centering
    \includegraphics[width=0.475\linewidth]{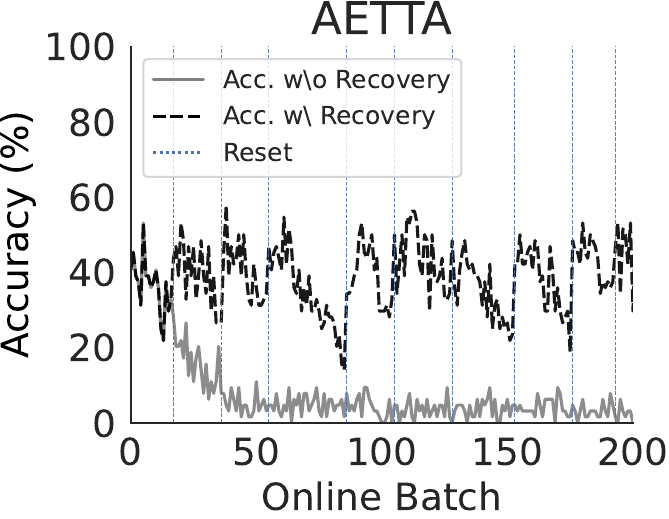}
    \includegraphics[width=0.475\linewidth]{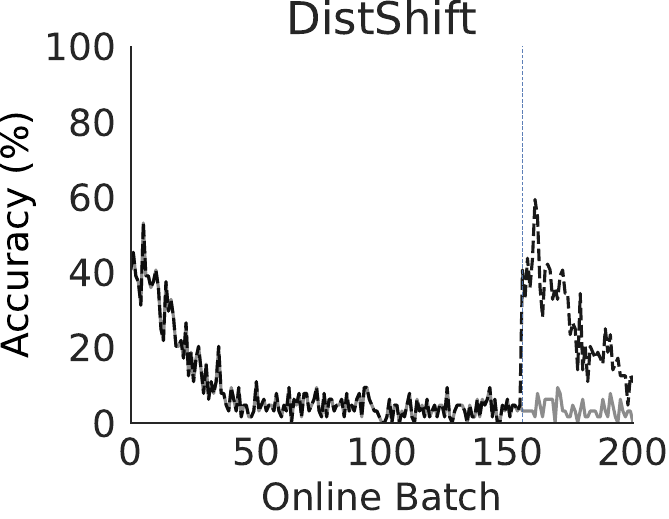}
    \end{subfigure}
  \caption{An example of model recovery compared with DistShift. Reset points are marked over the x-axis.
  }\label{fig:case_study}
\end{figure}

\noindent\textbf{Qualitative Analysis.}
Figure~\ref{fig:case_study} shows an example of our model recovery compared with DistShift. Notably, our recovery algorithm resets only when an accuracy degradation trend is detected. On the other hand, DistShift failed to recover in the early steps since it resets the model only on distribution shifts. This implies that estimating performance degradation is more beneficial than knowing when the domain changes to improve TTA performance. 

\section{Related Work}\label{sec:related_work}

\noindent\textbf{Test-Time Adaptation.} 
Recent progress in the field of test-time adaptation (TTA) has focused on improving model robustness~\cite{sar, note, rotta, cotta, lame, eata, sotta} and addressing novel forms of domain shifts~\cite{note, cotta, sotta}. 
On the other hand, an analysis~\cite{rdumb} pointed out the conventional TTA approaches remain prone to adaptation failures and demonstrated the importance of model recovery. In alignment with this insight, our work not only showcases the feasibility of accuracy estimation for TTA but also investigates a promising model recovery solution to enhance the robustness of TTA.\\

\noindent\textbf{Accuracy Estimation.} 
Existing accuracy estimation approaches mainly focus on the ensemble of multiple pre-trained models~\cite{accuracy_on_the_line, agreement_on_the_line, disagreement, DoC, chen2021ensemble}.
Accuracy-on-the-line~\cite{accuracy_on_the_line} and Agreement-on-the-line~\cite{agreement_on_the_line} have demonstrated a notable linear relationship between performances in a wide range of models and distribution shifts, relying on the consistency of model predictions between in-distribution (ID) and out-of-distribution (OOD) data. 
The Difference of Confidence (DoC)~\cite{DoC} leverages differences in the model's confidence between ID and OOD data to estimate the accuracy gap under distribution shifts for calculating the final OOD accuracy.
Self-training ensemble~\cite{chen2021ensemble} estimates the accuracy of the pre-trained classifier by iteratively learning an ensemble of models with a training dataset, unlabeled test dataset, and wrongly classified samples.
All these methods require labeled ID data to estimate OOD accuracy. To our knowledge, no existing studies target the accuracy estimation in TTA where source data and labels are unavailable.

\section{Conclusion}\label{sec:discussion}

We proposed a label-free TTA performance estimation method without access to source data and target labels. Based on the dropout inference sampling, we proposed calculating the prediction disagreement to estimate the TTA accuracy. We further improved the method with robust disagreement equality by utilizing the batch-aggregated distribution to penalize skewed predictions. Our method outperformed the baselines in diverse scenarios and datasets. Finally, our case study of model recovery showed the practicality of accuracy estimation. Our findings suggest that accuracy estimation is not only feasible but also a valuable tool in advancing the field of TTA without the need for labeled data.

\section*{Acknowledgements}

This work was supported by the Institute of Information \& communications Technology Planning \& Evaluation (IITP) grant funded by the Korea government (MSIT) (No.2022-0-00495, On-Device Voice Phishing Call Detection).

{
    \small
    \bibliographystyle{ieeenat_fullname}
    \bibliography{main}
}


\clearpage
\maketitlesupplementary
\appendix

\section{Proof of Theorems}

\subsection{Proof of Theorem~\ref{thm:class-wise}}\label{sec:app-class-wise}
We start expanding test error ${\tt Err}$ with few modifications from GDE~\cite{disagreement}:
\begin{flalign}
    &\mathbb{E}_{h \sim {\mathcal{H}}_{\mathcal{A}}} [{\tt Err}_{\mathcal{D^T}}(h)]&\\
    &\triangleq  \mathbb{E}_{{\mathcal{H}}_{\mathcal{A}}} [\mathbb{E}_{\mathcal{D^T}} [ \mathbb{1} ( h({X}; \Theta) \neq {Y} ) ]] & \\
    &= \mathbb{E}_{\mathcal{D^T}} [ \mathbb{E}_{{\mathcal{H}}_{\mathcal{A}}} [\mathbb{1} ( h({X}; \Theta) \neq {Y} )]  & \text{(exchanging expectations)} \\
    &= \mathbb{E}_{\mathcal{D^T}} [ 1 - \Tilde{h}_{{Y}} ({X}) ]\\
    &= \sum^{K-1}_{k=0} \int_{\mathbf{x}}^{} (1 - \Tilde{h}_k(\mathbf{x})) \; p({X} = \mathbf{x}, {Y} = k) d\mathbf{x} & (\text{by definition of expectation}) \\
    &= \int_{\boldsymbol{q} \in \Delta^K} \sum^{K-1}_{k=0} \int_{\mathbf{x}}^{} (1 - \Tilde{h}_k(\mathbf{x})) \; p({X} = \mathbf{x}, {Y} = k, \Tilde{h}({X}) = \boldsymbol{q}) d\mathbf{x} d\boldsymbol{q} & (\text{introducing} \; \Tilde{h} \; \text{as a r.v.}) \\
    &= \int_{\boldsymbol{q} \in \Delta^K} \sum^{K-1}_{k=0} \int_{\mathbf{x}}^{} (1 - \Tilde{h}_k(\mathbf{x})) \; p({Y} = k, \Tilde{h}({X}) = \boldsymbol{q}) p({X} = \mathbf{x} | {Y} = k, \Tilde{h}({X}) = \boldsymbol{q}) d\mathbf{x} d\boldsymbol{q} & \\
    &= \int_{\boldsymbol{q} \in \Delta^K} \sum^{K-1}_{k=0} p({Y} = k, \Tilde{h}({X}) = \boldsymbol{q}) \int_{\mathbf{x}}^{} (1 - \underbrace{\Tilde{h}_k(\mathbf{x})}_{= q_k}) \; p({X} = \mathbf{x} | {Y} = k, \Tilde{h}({X}) = \boldsymbol{q}) d\mathbf{x} d\boldsymbol{q} & \\
    &= \int_{\boldsymbol{q} \in \Delta^K} \sum^{K-1}_{k=0} p({Y} = k, \Tilde{h}({X}) = \boldsymbol{q}) \int_{\mathbf{x}}^{} \underbrace{(1 - q_k)}_{\text{constant w.r.t.} \int_{\mathbf{x}}} \; p({X} = \mathbf{x} | {Y} = k, \Tilde{h}({X}) = \boldsymbol{q}) d\mathbf{x} d\boldsymbol{q} & \\
    &= \int_{\boldsymbol{q} \in \Delta^K} \sum^{K-1}_{k=0} p({Y} = k, \Tilde{h}({X}) = \boldsymbol{q}) (1 - q_k) \underbrace{\int_{\mathbf{x}}^{} \; p({X} = \mathbf{x} | {Y} = k, \Tilde{h}({X}) = \boldsymbol{q}) d\mathbf{x}}_{=1} d\boldsymbol{q} & \\
    &= \int_{\boldsymbol{q} \in \Delta^K} \sum^{K-1}_{k=0} p({Y} = k, \Tilde{h}({X}) = \boldsymbol{q}) (1 - q_k) d\boldsymbol{q} & \\
    &= \int_{q \in [0, 1]} \sum^{K-1}_{k=0} p({Y} = k, \Tilde{h}_k({X}) = q) (1 - q) dq & (\text{refer \cite{disagreement}}) \\
    &= \int_{q \in [0, 1]} \sum^{K-1}_{k=0} \underbrace{p({Y} = k | \Tilde{h}_k({X}) = q)}_{= q} p(\Tilde{h}_k({X}) = q) (1 - q) dq & \\
    &= \int_{q \in [0, 1]} q (1 - q) \sum^{K-1}_{k=0} p(\Tilde{h}_k({X}) = q) dq.& (\text{confidence-prediction calibration})~\label{eq:err}
\end{flalign}
Then, we expand the prediction disagreement with dropouts (${\tt PDD}$) from its definition:
\begin{flalign}
    &\mathbb{E}_{h \sim {\mathcal{H}}_{\mathcal{A}}} [{\tt PDD}_{\mathcal{D^T}}(h)]&\\
    &\triangleq  \mathbb{E}_{{\mathcal{H}}_{\mathcal{A}}} \left[\mathbb{E}_{\mathcal{D^T}} \left[ \frac{1}{N} \sum_{i=1}^N \mathbb{1} [ h({X}; \Theta) \neq h({X}; \Theta^{{\tt dropout}_i})  ]  \right] \right]&\\
    &= \mathbb{E}_{\mathcal{D^T}} \left[\mathbb{E}_{{\mathcal{H}}_{\mathcal{A}}} \left[ \frac{1}{N} \sum_{i=1}^N \mathbb{1} [ h({X}; \Theta) \neq h({X}; \Theta^{{\tt dropout}_i})  ]  \right] \right]&(\text{exchanging expectations})\\
    &= \mathbb{E}_{\mathcal{D^T}} \left[ \mathbb{E}_{{\mathcal{H}}_{\mathcal{A}}} \left[ \frac{1}{N} \sum_{i=1}^N \sum_{k=0}^{K-1} \mathbb{1} [h({X}; \Theta) = k] (1 - \mathbb{1} [h({X}; \Theta^{{\tt dropout}_i}) = k]) \right] \right] &\\
    &= \mathbb{E}_{\mathcal{D^T}} \left[ \sum_{k=0}^{K-1} \frac{1}{N} \sum_{i=1}^N \mathbb{E}_{{\mathcal{H}}_{\mathcal{A}}} \left[ \mathbb{1} [h({X}; \Theta) = k] (1 - \mathbb{1} [h({X}; \Theta^{{\tt dropout}_i}) = k]) \right] \right] &\\
    &= \mathbb{E}_{\mathcal{D^T}} \left[ \sum_{k=0}^{K-1} \mathbb{E}_{{\mathcal{H}}_{\mathcal{A}}}  \left[\mathbb{1} [h({X}; \Theta) = k] \right] \mathbb{E}_{{\mathcal{H}}_{\mathcal{A}}} [1 - \frac{1}{N} \sum_{i=1}^N \mathbb{1} [h({X}; \Theta^{{\tt dropout}_i}) = k]]  \right] & (\text{Dropout independence (Definition~\ref{def:efa})})\\
    &= \mathbb{E}_{\mathcal{D^T}} \left[ \sum_{k=0}^{K-1} \Tilde{h}_k ({X}) (1 - \Tilde{h}_k ({X}) ) \right]&\\
    &= \int_{\mathbf{x}} \sum_{k=0}^{K-1} \Tilde{h}_k (\mathbf{x}) (1 - \Tilde{h}_k (\mathbf{x}))  p({X} = \mathbf{x}) d\mathbf{x} & (\text{by definition of expectation})\\
    &= \int_{\boldsymbol{q} \in \Delta^K} \int_{\mathbf{x}}^{} \sum^{K-1}_{k=0} {\Tilde{h}_k (\mathbf{x}) (1 - \Tilde{h}_k (\mathbf{x}))}  p\left({X} = \mathbf{x}, \Tilde{h} ({X}) = \boldsymbol{q} \right) d\mathbf{x} d\boldsymbol{q} &(\text{introducing} \; \Tilde{h} \; \text{as a r.v.})\\
    &= \int_{\boldsymbol{q} \in \Delta^K} p (\Tilde{h} ({X}) = \boldsymbol{q}) \int_{\mathbf{x}}^{} \sum^{K-1}_{k=0} \underbrace{\Tilde{h}_k (\mathbf{x}) (1 - \Tilde{h}_k (\mathbf{x}))}_{\Tilde{h}_k (\mathbf{x}) = q_k} p\left( {X} = \mathbf{x} | \Tilde{h} ({X}) = \boldsymbol{q} \right) d\mathbf{x} d\boldsymbol{q} & \\
    &= \int_{\boldsymbol{q} \in \Delta^K} p (\Tilde{h} ({X}) = \boldsymbol{q}) \int_{\mathbf{x}}^{} \underbrace{\sum^{K-1}_{k=0}}_{\text{bring to the front}} q_k (1 - q_k) p( {X} = \mathbf{x} | \Tilde{h} ({X}) = \boldsymbol{q} ) d\mathbf{x} d\boldsymbol{q} &\\
    &= \sum^{K-1}_{k=0} \int_{\boldsymbol{q} \in \Delta^K} p (\Tilde{h} ({X}) = \boldsymbol{q}) \int_{\mathbf{x}}^{} \underbrace{q_k (1 - q_k)}_{\text{constant w.r.t.} \; \int_{\mathbf{x}}} p( {X} = \mathbf{x} | \Tilde{h} ({X}) = \boldsymbol{q} ) d\mathbf{x} d\boldsymbol{q} &\\
    &= \underbrace{\sum^{K-1}_{k=0} \int_{\boldsymbol{q} \in \Delta^K}}_{\text{swap}} p (\Tilde{h} ({X}) = \boldsymbol{q}) q_k (1 - q_k) \underbrace{\int_{\mathbf{x}}^{} p( {X} = \mathbf{x} | \Tilde{h} ({X}) = \boldsymbol{q} ) d\mathbf{x}}_{=1} d\boldsymbol{q} &\\
    &= \int_{\boldsymbol{q} \in \Delta^K} \sum^{K-1}_{k=0} q_k (1 - q_k) p (\Tilde{h} ({X}) = \boldsymbol{q}) d\boldsymbol{q} &\\
    &= \int_{q \in [0, 1]} q (1 - q) \sum^{K-1}_{k=0} p (\Tilde{h}_k ({X}) = q) dq. & (\text{refer~\cite{disagreement}})~\label{eq:pda}
\end{flalign}
Equation~\ref{eq:err} is equivalent to Equation~\ref{eq:pda}:
\begin{flalign}
    & \mathbb{E}_{h \sim {\mathcal{H}}_{\mathcal{A}}} [{\tt Err}_{\mathcal{D^T}}(h)] = 
    \mathbb{E}_{h \sim {\mathcal{H}}_{\mathcal{A}}} [{\tt PDD}_{\mathcal{D^T}}(h)],&
\end{flalign}
which concludes the proof of Theorem~\ref{thm:class-wise}.

\subsection{Proof of Theorem~\ref{thm:shifted-class-wise}}\label{sec:app-shifted-class-wise}
From robust confidence-prediction calibration, the over-confident model's conditional probability of the major class $k'$ is scaled by $a$, while other classes' conditional probabilities are equally scaled up by $b$.
Then, Equation~\ref{eq:err} now becomes:
\begin{flalign}
    &\mathbb{E}_{h \sim {\mathcal{H}}_{\mathcal{A}}} [{\tt Err}_{\mathcal{D^T}}(h)]  &\\
    &= \int_{q \in [0, 1]} \sum^{K-1}_{k=0} {p({Y} = k | \Tilde{h}_k({X}) = q)} p(\Tilde{h}_k({X}) = q) (1 - q) dq & \\
    &= \int_{q \in [0, 1]} {p({Y} = k' | \Tilde{h}_{k'}({X}) = q)} p(\Tilde{h}_{k'}({X}) = q) (1 - q) + \sum^{}_{k \neq k'} {p({Y} = k | \Tilde{h}_k({X}) = q)} p(\Tilde{h}_k({X}) = q) (1 - q) dq & \\ 
    &= \int_{q \in [0, 1]} a q \; p(\Tilde{h}_{k'}({X}) = q) (1 - q) + \sum^{}_{k \neq k'} b q \; p(\Tilde{h}_k({X}) = q) (1 - q) dq & \nonumber \\ 
    & \qquad\qquad\qquad\qquad\qquad\qquad\qquad\qquad\qquad\qquad\qquad\qquad\qquad\qquad\;\; (\text{robust confidence-prediction calibration}) \\
    &= \int_{q \in [0, 1]} a q (1 - q) \; p(\Tilde{h}_{k'}({X}) = q) dq \; + \; b \int_{q \in [0, 1]} \sum^{}_{k \neq k'} q (1 - q) \; p(\Tilde{h}_k({X}) = q) dq. &~\label{eq:rewrite}
\end{flalign}
We rewrite Equation~\ref{eq:rewrite} as:
\begin{flalign}
    & \int_{q \in [0, 1]} \sum^{}_{k \neq k'} q (1 - q) \; p(\Tilde{h}_k({X}) = q) dq = \frac{1}{b} \mathbb{E}_{h \sim {\mathcal{H}}_{\mathcal{A}}} [{\tt Err}_{\mathcal{D^T}}(h)] - \int_{q \in [0, 1]} \frac{a}{b} \; q (1 - q) \; p(\Tilde{h}_{k'}({X}) = q) dq . &~\label{eq:subst}
\end{flalign}
Then, we rewrite PDD (Equation~\ref{eq:pda}):
\begin{flalign}
    &\mathbb{E}_{h \sim {\mathcal{H}}_{\mathcal{A}}} [{\tt PDD}_{\mathcal{D^T}}(h)] &\\
    &= \int_{q \in [0, 1]} q (1 - q) \sum^{K-1}_{k=0} p (\Tilde{h}_k ({X}) = q) dq&\\
    &= \int_{q \in [0, 1]} q (1 - q) \; p (\Tilde{h}_{k'} ({X}) = q) + q (1 - q) \sum^{}_{k \neq k'} p (\Tilde{h}_k ({X}) = q) dq&\\
    &= \int_{q \in [0, 1]} q (1 - q) \; p (\Tilde{h}_{k'} ({X}) = q) dq \; + \; \int_{q \in [0, 1]} q (1 - q) \sum^{}_{k \neq k'} p (\Tilde{h}_k ({X}) = q) dq&\\
    &= \int_{q \in [0, 1]} \frac{b-a}{b} \; q (1 - q) \; p (\Tilde{h}_{k'} ({X}) = q) dq \; + \; \frac{1}{b} \mathbb{E}_{h \sim {\mathcal{H}}_{\mathcal{A}}} [{\tt Err}_{\mathcal{D^T}}(h)]. & (\text{Equation~\ref{eq:subst}})
\end{flalign}
Finally, we obtain the equality between $\tt Err$ and $\tt PDD$:
\begin{flalign}
    & \mathbb{E}_{h \sim {\mathcal{H}}_{\mathcal{A}}} [{\tt Err}_{\mathcal{D^T}}(h)] & \\
    & = b \; \mathbb{E}_{h \sim {\mathcal{H}}_{\mathcal{A}}} [{\tt PDD}_{\mathcal{D^T}}(h)] - \int_{q \in [0, 1]} {(b-a)} \; q (1 - q) \; p (\Tilde{h}_{k'} ({X}) = q) dq, & 
\end{flalign}
which concludes the proof of Theorem~\ref{thm:shifted-class-wise}. Note that without weighting ($a=b=1$), the result is identical to Theorem~\ref{thm:class-wise}.

\newpage
\section{Additional Experiments}\label{sec:app-add-exp}

\paragraph{GDE with multiple pre-trained models.} We compare \system{} with the original version of GDE (denoted as GDE*), utilizing multiple pre-trained models with access to training data. We report the result in Table~\ref{tab:supp_gde}. Due to the misalignment of confidence-prediction calibration, GDE* underperforms \system{} even with full access to source data.

\begin{table*}[ht]
\centering
\caption{Mean absolute error (MAE) (\%) of the accuracy estimation on continual CIFAR100-C.}
\label{tab:supp_gde}
\smaller
\begin{tabularx}{\textwidth}{ll*{6}Y>{\columncolor[HTML]{EDEEFF}}Y}
\Xhline{2\arrayrulewidth}
\addlinespace[0.08cm] 
 &  & \multicolumn{6}{c}{TTA Method} & \cellcolor[HTML]{FFFFFF} \\ \cline{3-8}
\addlinespace[0.08cm] 
\multirow{-2}{*}{Dataset} & \multirow{-2}{*}{Method} & TENT~\cite{tent} & EATA~\cite{eata} & SAR~\cite{sar} & CoTTA~\cite{cotta} & RoTTA~\cite{rotta} & SoTTA~\cite{sotta} & {\cellcolor[HTML]{EDEEFF}Avg. ($\downarrow$)} \\ \hline
& GDE*~\cite{disagreement} & 14.54 \scalebox{\std}{± 8.14} & 4.11 \scalebox{\std}{± 2.43} & 7.27 \scalebox{\std}{± 0.16} & 9.89 \scalebox{\std}{± 0.29} & 7.44 \scalebox{\std}{± 0.13} & 5.79 \scalebox{\std}{± 0.23} & 8.17 \scalebox{\std}{± 0.87} \\
\multirow{-2}{*}{\begin{tabular}[c]{@{}l@{}}Continual\\ CIFAR100-C\end{tabular}} & \cellcolor[HTML]{E2FFE1}\system{} & \cellcolor[HTML]{E2FFE1}5.85 \scalebox{\std}{± 0.36} & \cellcolor[HTML]{E2FFE1}4.18 \scalebox{\std}{± 0.82} & \cellcolor[HTML]{E2FFE1}6.67 \scalebox{\std}{± 0.12} & \cellcolor[HTML]{E2FFE1}6.55 \scalebox{\std}{± 0.17} & \cellcolor[HTML]{E2FFE1}5.86 \scalebox{\std}{± 0.10} & \cellcolor[HTML]{E2FFE1}5.32 \scalebox{\std}{± 0.18} & \cellcolor[HTML]{E2FFE1}\textbf{5.74 \scalebox{\std}{± 0.13}} \\ 
\Xhline{2\arrayrulewidth}
\end{tabularx}
\end{table*}

\paragraph{ImageNet-R.} To demonstrate the dataset generality of \system{}, we report the accuracy estimation result on ResNet18 architecture on ImageNet-R (Table~\ref{tab:imagenetR}). \system{} outperformed the baselines in all TTA methods, showing \system{} is applicable in various datasets (e.g., CIFAR10-C, CIFAR100-C, ImageNet-C, and ImageNet-R).

\begin{table*}[ht]
\centering
\caption{Mean absolute error (MAE) (\%) of the accuracy estimation on ResNet18 on ImageNet-R. \textbf{Bold} numbers are the lowest error. Averaged over three different random seeds.}

\label{tab:imagenetR}
\smaller
\begin{tabularx}{\textwidth}{ll*{6}Y>{\columncolor[HTML]{EDEEFF}}Y}
\Xhline{2\arrayrulewidth}
\addlinespace[0.08cm] 
 &  & \multicolumn{6}{c}{TTA Method} & \cellcolor[HTML]{FFFFFF} \\ \cline{3-8}
\addlinespace[0.08cm] 
\multirow{-2}{*}{Dataset} & \multirow{-2}{*}{Method} & TENT~\cite{tent} & EATA~\cite{eata} & SAR~\cite{sar} & CoTTA~\cite{cotta} & RoTTA~\cite{rotta} & SoTTA~\cite{sotta} & {\cellcolor[HTML]{EDEEFF}Avg. ($\downarrow$)} \\ \hline
 & SrcValid & 37.00 \scalebox{\std}{± 0.14} & 37.91 \scalebox{\std}{± 0.18} & 36.58 \scalebox{\std}{± 0.24} & 35.05 \scalebox{\std}{± 0.16} & 34.43 \scalebox{\std}{± 0.05} & 37.82 \scalebox{\std}{± 0.41} & 36.46 \scalebox{\std}{± 0.05}
 \\
 & SoftmaxScore~\cite{softmaxscore} & 10.79 \scalebox{\std}{± 0.17} & 13.87 \scalebox{\std}{± 0.09} & 15.02 \scalebox{\std}{± 0.14} & 14.76 \scalebox{\std}{± 0.07} & 14.08 \scalebox{\std}{± 0.04} & 12.25 \scalebox{\std}{± 0.42} & 13.46 \scalebox{\std}{± 0.06}
 \\
 & GDE~\cite{disagreement} & 62.81 \scalebox{\std}{± 0.12} & 61.36 \scalebox{\std}{± 0.14} & 63.27 \scalebox{\std}{± 0.18} & 64.86 \scalebox{\std}{± 0.10} & 62.64 \scalebox{\std}{± 0.12} & 55.23 \scalebox{\std}{± 0.29} & 61.70 \scalebox{\std}{± 0.02}
 \\
 & AdvPerturb~\cite{advperturb} & 13.42 \scalebox{\std}{± 0.28} & 16.04 \scalebox{\std}{± 0.36} & 17.90 \scalebox{\std}{± 0.28} & 21.19 \scalebox{\std}{± 0.22} & 31.12 \scalebox{\std}{± 0.12} & 9.91 \scalebox{\std}{± 0.73} & 18.26 \scalebox{\std}{± 0.03}
 \\
\multirow{-5}{*}{ImageNet-R} & \cellcolor[HTML]{E2FFE1}\system{} & \cellcolor[HTML]{E2FFE1}8.02 \scalebox{\std}{± 0.12} & \cellcolor[HTML]{E2FFE1}6.87 \scalebox{\std}{± 0.08} & \cellcolor[HTML]{E2FFE1}7.06 \scalebox{\std}{± 0.19} & \cellcolor[HTML]{E2FFE1}7.07 \scalebox{\std}{± 0.11} & \cellcolor[HTML]{E2FFE1}8.63 \scalebox{\std}{± 0.19} & \cellcolor[HTML]{E2FFE1}6.79 \scalebox{\std}{± 0.29} & \textbf{\cellcolor[HTML]{E2FFE1}7.41 \scalebox{\std}{± 0.05}}
\\
\Xhline{2\arrayrulewidth}
\end{tabularx}
\end{table*}

\paragraph{ResNet50.} To demonstrate the model generality of \system{}, we report the accuracy estimation result on ResNet50 architecture on ImageNet-C (Table~\ref{tab:resnet50}). \system{} outperformed the baselines in general, showing \system{} is applicable to diverse model architectures.

\begin{table*}[ht]
\centering
\caption{Mean absolute error (MAE) (\%) of the accuracy estimation on ResNet50 on ImageNet-C. \textbf{Bold} numbers are the lowest error. Averaged over three different random seeds for 15 types of corruption.}

\label{tab:resnet50}
\smaller
\begin{tabularx}{\textwidth}{ll*{6}Y>{\columncolor[HTML]{EDEEFF}}Y}
\Xhline{2\arrayrulewidth}
\addlinespace[0.08cm] 
 &  & \multicolumn{6}{c}{TTA Method} & \cellcolor[HTML]{FFFFFF} \\ \cline{3-8}
\addlinespace[0.08cm] 
\multirow{-2}{*}{Dataset} & \multirow{-2}{*}{Method} & TENT~\cite{tent} & EATA~\cite{eata} & SAR~\cite{sar} & CoTTA~\cite{cotta} & RoTTA~\cite{rotta} & SoTTA~\cite{sotta} & {\cellcolor[HTML]{EDEEFF}Avg. ($\downarrow$)} \\ \hline
 & SrcValid & 46.46 \scalebox{\std}{± 0.15} & 34.19 \scalebox{\std}{± 0.67} & 30.35 \scalebox{\std}{± 0.75} & 46.47 \scalebox{\std}{± 0.17} & 12.28 \scalebox{\std}{± 0.11} & 19.28 \scalebox{\std}{± 0.18} & 31.50 \scalebox{\std}{± 0.26} \\
 & SoftmaxScore~\cite{softmaxscore} & 23.58 \scalebox{\std}{± 0.03} & 24.14 \scalebox{\std}{± 0.04} & 26.22 \scalebox{\std}{± 0.05} & 23.57 \scalebox{\std}{± 0.04} & 24.32 \scalebox{\std}{± 0.02} & 17.87 \scalebox{\std}{± 0.24} & 23.28 \scalebox{\std}{± 0.03} \\
 & GDE~\cite{disagreement} & 68.39 \scalebox{\std}{± 0.04} & 57.08 \scalebox{\std}{± 0.08} & 55.81 \scalebox{\std}{± 0.06} & 68.36 \scalebox{\std}{± 0.04} & 58.69 \scalebox{\std}{± 0.09} & 48.15 \scalebox{\std}{± 0.33} & 59.41 \scalebox{\std}{± 0.04} \\
 & AdvPerturb~\cite{advperturb} & 12.77 \scalebox{\std}{± 0.04} & 21.16 \scalebox{\std}{± 0.05} & 23.66 \scalebox{\std}{± 0.08} & 12.77 \scalebox{\std}{± 0.05} & 16.44 \scalebox{\std}{± 0.00} & 25.28 \scalebox{\std}{± 0.32} & 18.68 \scalebox{\std}{± 0.05} \\
\multirow{-5}{*}{\begin{tabular}[c]{@{}l@{}}Fully\\ ImageNet-C\end{tabular}} & \cellcolor[HTML]{E2FFE1}\system{} & \cellcolor[HTML]{E2FFE1}6.14 \scalebox{\std}{± 0.05} & \cellcolor[HTML]{E2FFE1}9.15 \scalebox{\std}{± 0.03} & \cellcolor[HTML]{E2FFE1}8.50 \scalebox{\std}{± 0.07} & \cellcolor[HTML]{E2FFE1}6.15 \scalebox{\std}{± 0.04} & \cellcolor[HTML]{E2FFE1}28.28 \scalebox{\std}{± 0.03} & \cellcolor[HTML]{E2FFE1}36.90 \scalebox{\std}{± 0.36} & \cellcolor[HTML]{E2FFE1}\textbf{15.85 \scalebox{\std}{± 0.09}} \\ \hline
 & SrcValid & 46.38 \scalebox{\std}{± 0.10} & 35.83 \scalebox{\std}{± 0.74} & 24.35 \scalebox{\std}{± 1.86} & 46.46 \scalebox{\std}{± 0.22} & 13.79 \scalebox{\std}{± 0.16} & 5.12 \scalebox{\std}{± 0.29} & 28.65 \scalebox{\std}{± 0.46} \\
 & SoftmaxScore~\cite{softmaxscore} & 23.58 \scalebox{\std}{± 0.03} & 21.34 \scalebox{\std}{± 0.06} & 16.64 \scalebox{\std}{± 0.25} & 23.61 \scalebox{\std}{± 0.01} & 19.99 \scalebox{\std}{± 0.25} & 51.60 \scalebox{\std}{± 0.75} & 26.13 \scalebox{\std}{± 0.12} \\
 & GDE~\cite{disagreement} & 68.36 \scalebox{\std}{± 0.03} & 58.41 \scalebox{\std}{± 0.14} & 60.20 \scalebox{\std}{± 0.24} & 68.38 \scalebox{\std}{± 0.01} & 68.98 \scalebox{\std}{± 0.52} & 86.08 \scalebox{\std}{± 0.36} & 68.40 \scalebox{\std}{± 0.09} \\
 & AdvPerturb~\cite{advperturb} & 12.80 \scalebox{\std}{± 0.04} & 19.82 \scalebox{\std}{± 0.12} & 21.50 \scalebox{\std}{± 0.14} & 12.77 \scalebox{\std}{± 0.02} & 13.77 \scalebox{\std}{± 0.35} & 4.79 \scalebox{\std}{± 0.17} & 14.24 \scalebox{\std}{± 0.07} \\
\multirow{-5}{*}{\begin{tabular}[c]{@{}l@{}}Continual\\ ImageNet-C\end{tabular}} & \cellcolor[HTML]{E2FFE1}\system{} & \cellcolor[HTML]{E2FFE1}6.15 \scalebox{\std}{± 0.05} & \cellcolor[HTML]{E2FFE1}10.81 \scalebox{\std}{± 0.01} & \cellcolor[HTML]{E2FFE1}6.41 \scalebox{\std}{± 0.08} & \cellcolor[HTML]{E2FFE1}6.00 \scalebox{\std}{± 0.04} & \cellcolor[HTML]{E2FFE1}14.90 \scalebox{\std}{± 0.30} & \cellcolor[HTML]{E2FFE1}4.21 \scalebox{\std}{± 0.12} & \cellcolor[HTML]{E2FFE1}\textbf{8.08 \scalebox{\std}{± 0.04}}
\\
\Xhline{2\arrayrulewidth}
\end{tabularx}
\end{table*}

\newpage

\section{Discussion}

\paragraph{Potential Societal Impact.}  The computational overheads associated with test-time adaptation (TTA) could raise environmental concerns, particularly regarding carbon emissions. Our algorithm introduces $N$ extra model inferences for accuracy estimation. Importantly, our approach of utilizing dropout inference is computationally lightweight compared to baseline methods involving model retraining~\cite{disagreement} and adversarial backpropagation~\cite{advperturb}. Recent advancements, such as the memory-economic TTA~\cite{hong2023mecta}, are anticipated to tackle these challenges effectively. This implies that, despite the computational demands, the environmental impact of our approach could be mitigated by integrating emerging strategies for resource-efficient TTA implementations.

\paragraph{Limitations and Future Directions.} Our research investigates the possibility of accuracy estimation for TTA with only unlabeled data. A promising direction for further improvements is the (1) optimization of the weighting constant $b$ (or corresponding $a$), which stands to fine-tune the calibration process, and (2) estimation of the variable $C$ for more precise error estimates.
Also, we presented a case study on model recovery to demonstrate the practicality of accuracy estimation. While we chose a heuristic method to reset the model for the simplicity of analysis, there exists room for improvement to be more effective. Beyond model recovery, we also envision the potential of accuracy estimation in broader applications, such as model refinement and maintenance processes, and enhancing the dynamics of human-AI interactions, which we leave as future work.

\section{Experiment Details}\label{sec:app-exp-details}

We conducted all experiments under three random seeds (0, 1, 2) and reported the average values with standard deviations. The experiments were performed on NVIDIA GeForce RTX 3090 and NVIDIA TITAN RTX GPUs.

\subsection{Accuracy Estimation Details}\label{sec:app-acc-est-details}

\paragraph{\system{} (Ours).} We used the number of dropout inference samples $N=10$ and   prediction disagreement weighting hyperparameter $\alpha=3$ for all experiments. The maximum entropy for the model $E^{\tt{max}}$ is calculated as $E^{\tt{max}} = {\tt Ent} (\vec{1}_K / K)$ where $K$ is a number of classes and $\vec{1}_K$ is one vector with size $K$; which results in 2.3, 4.6, {and} 6.9 for 10, 100, {and} 1,000 classes. We applied the Dropout module for each residual block layer in ResNet18~\cite{7780459}, where the dropout rate is 0.4, 0.3, {and} 0.2 for 10, 100, {and} 1,000 classes, following previous studies which apply different hyperparameters for different numbers of classes~\cite{eata, cotta, sotta}.

\paragraph{SrcValid.} For SrcValid, we used labeled source-domain validation data and calculated the accuracy. We used 1,000 random samples from the validation set of the source dataset.

\paragraph{SoftmaxScore.} For SoftmaxScore~\cite{softmaxscore}, we utilized the average softmax score for the current test batch as the estimated accuracy. We additionally applied temperature scaling~\cite{guo2017calibration} with temperature value $T=2$, which showed the best estimation performance on CIFAR10-C, to enhance the estimation performance.

\paragraph{GDE.} For generalization disagreement equality (GDE)~\cite{disagreement}, we calculated the (dis)agreement rate between predictions of the test batch over a pair of models. Unlike the setting in domain adaptation of utilizing multiple pre-trained models, we utilized the models in different adaptation stages. Specifically, we compared the two models: (1) the currently adapted model and (2) the previous model right before the adaptation. This follows the suggestion that utilizing only two models is sufficient to calculate disagreement~\cite{disagreement}.

\paragraph{AdvPerturb.} Adversarial perturbation~\cite{advperturb} estimates the source model accuracy by calculating the agreement between the domain-adapted and source models by applying adversarial perturbation on the source model side. In the TTA setting, we compared the test-time-adapted model with the source model and applied the FGSM~\cite{fgsm} adversarial attack with attack size following the original paper ($\epsilon = 1/255$).

\subsection{TTA Method Details}\label{sec:app-tta-details}
In this study, we followed the official implementation of TTA methods.
To maintain consistency, we adopted the optimal hyperparameters reported in the corresponding papers or source code repositories. We also provide additional implementation details and the use of hyperparameters if not specified in the original paper or the source code.

\paragraph{TENT.} For TENT~\cite{tent}, we configured the learning rate as $LR = 0.001$ for CIFAR10-C/CIFAR100-C and $LR = 0.00025$ for ImageNet-C, aligning with the guidelines outlined in the original paper. The implementation followed the official code.\footnote{\url{https://github.com/DequanWang/tent}}

\paragraph{EATA.} For EATA~\cite{eata}, we followed the original configuration of $LR = 0.005/0.005/0.00025$ for CIFAR10-C/CIFAR100-C/ImageNet-C, entropy constant $E_0 = 0.4 \times \ln K$, where $K$ represents the number of classes. Additionally, we set the cosine sample similarity threshold $\epsilon = 0.4/0.4/0.05$, trade-off parameter $\beta = 1/1/2,000$, and moving average factor $\alpha = 0.1$. The Fisher importance calculation involved 2,000 samples, as recommended. The implementation followed the official code.\footnote{\url{https://github.com/mr-eggplant/EATA}}

\paragraph{SAR.} For SAR~\cite{sar}, we selected a batch size 64 for fair comparisons. We set a learning rate of $LR=0.00025$, sharpness threshold $\rho = 0.5$, and entropy threshold $E_0 = 0.4 \times \mathrm{ln} K$, following the recommendations from the original paper. The top layer (layer 4 for ResNet18) was frozen, consistent with the original paper. The implementation followed the official code.\footnote{\url{https://github.com/mr-eggplant/SAR}}

\paragraph{CoTTA.} For CoTTA~\cite{cotta}, we set the restoration factor $p=0.01$, and exponential moving average (EMA) factor $\alpha=0.999$. For augmentation confidence threshold $p_{th}$, we followed the authors' guidelines as $p_{th}=0.92$ for CIFAR10-C, $p_{th}=0.72$ for CIFAR100-C, and $p_{th}=0.1$ for ImageNet-C. The implementation followed the official code.\footnote{\url{https://github.com/qinenergy/cotta}}

\paragraph{RoTTA.} For RoTTA~\cite{rotta}, we utilized the Adam optimizer~\cite{kingma:adam} with a learning rate of $LR = 0.001$ and $\beta = 0.9$. We followed the original hyperparameters, including BN-statistic exponential moving average updating rate $\alpha = 0.05$, Teacher model's exponential moving average updating rate $\nu = 0.001$, timeliness parameter $\lambda_t = 1.0$, and uncertainty parameter $\lambda_u = 1.0$. The implementation followed the original code.\footnote{\url{https://github.com/BIT-DA/RoTTA}}

\paragraph{SoTTA.} For SoTTA~\cite{sotta}, the Adam optimizer~\cite{kingma:adam} was employed, featuring a BN momentum of $m = 0.2$ and a learning rate of $LR = 0.001$ with a single adaptation epoch. The memory size was set to 64, with the confidence threshold $C_0$ configured as 0.99 for CIFAR10-C (10 classes), 0.66 for CIFAR100-C (100 classes), and 0.33 for ImageNet-C (1,000 classes). The entropy-sharpness L2-norm constraint $\rho$ was set to 0.5, aligning with the suggestion~\cite{sam}. The top layer was frozen following the original paper. 
The implementation followed the original code.\footnote{\url{https://github.com/taeckyung/sotta}}

\subsection{Experiment Setting Details}

\paragraph{Datasets.} CIFAR10-C/CIFAR100-C/ImageNet-C~\cite{cifarc} are the most widely used benchmarks for test-time adaptation (TTA)~\cite{tent, eata, sar, cotta, rotta, sotta, note}. All datasets contain 15 corruption types, including Gaussian, Snow, Frost, Fog, Brightness, Contrast, Elastic Transformation, Pixelate, and JPEG Compression. Each corruption is applied in 5 levels of severity, where we adopt the highest severity level of 5. CIFAR10-C and CIFAR100-C consist of 50,000 train images and 10,000 test images for 10 and 100 classes. ImageNet-C consists of 1,281,167 train images and 50,000 test images for 1,000 classes.

\paragraph{Pre-Training.} We employed the ResNet18~\cite{7780459} as the backbone network. The model is trained for each CIFAR10-C/CIFAR100-C/ImageNet-C on the training dataset. For CIFAR10-C/CIFAR100-C, we utilized the stochastic gradient descent with a batch size of 128, a learning rate of 0.1, and a momentum of 0.9, with cosine annealing learning rate scheduling~\cite{loshchilov2016sgdr} for 200 epochs. For ImageNet-C, we utilized the pre-trained model from TorchVision~\cite{torchvision2016}.

\paragraph{Test-Time Adaptation.} For the \textit{fully} TTA, each TTA method adapts to one corruption at a time. For the \textit{continual} TTA, each TTA method continually adapts to 15 corruptions in the predefined order of [Gaussian, Snow, Frost, Fog, Brightness, Contrast, Elastic Transformation, Pixelate, and JPEG Compression], following the previous study~\cite{cotta}. For all experiments, we use the batch size of 64, with memory size 64 for RoTTA~\cite{rotta} and SoTTA~\cite{sotta} for a fair comparison.

\subsection{Model Recovery Details (Section~\ref{sec:recovery})}

\paragraph{\system{} (Ours).} With \system{}, our reset algorithm detects two cases: (1) consecutive low accuracies and (2) sudden accuracy drops. For consecutive low accuracies, we utilize the information of estimated accuracy from each 5 batches. 
Regarding hard lower-bound thresholding, we employ a threshold value of 0.2. We reset both the model's weights to those from the source model and the optimizer's state to its initialization value.

\paragraph{Episodic.} Episodic resetting was first introduced by MEMO~\cite{memo}, where the model resets after every batch. We reset both the model's weights and the optimizer's state to its value before adaptation.

\paragraph{MRS.} The Model Recovery Scheme (MRS) was initially introduced by SAR ~\cite{sar} to recover the model from collapsing. The reset occurs when the moving average of entropy loss falls below a certain threshold. We utilized the threshold value of 0.2 introduced in the original paper. We reset both the model's weights to those from the source model and the optimizer's state to its initialization value.

\paragraph{Stochastic.} Stochastic restoration was first introduced by CoTTA ~\cite{cotta}. A small number of model weights are stochastically restored to the initial weights of the source model, with a certain probability specified by the restoration factor. We use the restoration factor 0.01, as introduced in the original work.

\paragraph{FisherStochastic.} Fisher information based restoration was proposed by PETAL ~\cite{petal}, based on the stochastic restoration~\cite{cotta}. It applies stochastic restoration based on layer importance measured by the Fisher information matrix (FIM). We use an FIM-based parameter restoration quantile value of 0.03 for CIFAR100-C, as recommended in the original paper. The parameter with an FIM value less than 0.03-quantile would be restored to the original source weight. 

\paragraph{DistShift.} DistShift assumes that the model knows when the distribution changes and thus acts as an oracle. Resetting occurs when the test data distribution (corruption) changes. We reset both the model's weights to those from the source model and the optimizer's state to its initialization value.

\section{License of Assets}\label{app:license}

\paragraph{Datasets.} CIFAR10/CIFAR100 (MIT License), CIFAR10-C/CIFAR100-C (Creative Commons Attribution 4.0 International) and ImageNet-C (Apache 2.0).

\paragraph{Codes.} Torchvision for ResNet18 and ResNet50 (Apache 2.0), the official repository of TENT (MIT License), the official repository of EATA (MIT License), the official repository of SAR (BSD 3-Clause License), the official repository of CoTTA (MIT License), the official repository of RoTTA (MIT License), and the official repository of SoTTA (MIT License).

\newpage

\section{Result Details}\label{sec:app-result-details}

We report the detailed results per corruption in the main experiments.  
Table~\ref{tab:main_experiment_iid} in the main paper is detailed in Table~\ref{tab:main_experiment_cifar10_full}, Table~\ref{tab:main_experiment_cifar100_full}, and Table~\ref{tab:main_experiment_imagenet_full}.
Table~\ref{tab:main_experiment_cont} in the main paper is detailed in Table~\ref{tab:main_experiment_cifar10_cont}, Table~\ref{tab:main_experiment_cifar100_cont}, and Table~\ref{tab:main_experiment_imagenet_cont}.
Table~\ref{tab:app_reset} in the main paper is detailed in Table~\ref{tab:model_recovery_all}.

\renewcommand\cellset{\renewcommand\arraystretch{0.8}%
\setlength\extrarowheight{0pt}}

\renewcommand{\arraystretch}{2.0}

\begin{table}[H]
\centering
\caption{Mean absolute error (MAE) (\%) of the accuracy estimation on fully CIFAR10-C. Averaged over three different random seeds.}
\label{tab:main_experiment_cifar10_full}
\scriptsize
\setlength\tabcolsep{3pt}
\begin{tabularx}{\linewidth}{ll*{16}Y}
\Xhline{2\arrayrulewidth}
 &  & \multicolumn{3}{c}{Noise} & \multicolumn{4}{c}{Blur} & \multicolumn{4}{c}{Weather} & \multicolumn{4}{c}{Digital} &  \\ 
\addlinespace[-0.05cm]
\cmidrule(lr){3-5} \cmidrule(lr){6-9} \cmidrule(lr){10-13} \cmidrule(lr){14-17}
\addlinespace[-0.05cm]
TTA Method & Acc. Estimation & Gau. & Shot & Imp. & Def. & Gla. & Mot. & Zoom & Snow & Fro. & Fog & Brit. & Cont. & Elas. & Pix. & JPEG & \cellcolor[HTML]{EDEEFF}Avg.($\downarrow$) \\ \midrule
  & SrcValid & \makecell[c]{24.85\\ \scalebox{\std}{± 0.83}} & \makecell[c]{21.82\\ \scalebox{\std}{± 0.74}} & \makecell[c]{32.41\\ \scalebox{\std}{± 0.90}} & \makecell[c]{11.60\\ \scalebox{\std}{± 0.53}} & \makecell[c]{32.49\\ \scalebox{\std}{± 1.50}} & \makecell[c]{12.78\\ \scalebox{\std}{± 0.45}} & \makecell[c]{11.16\\ \scalebox{\std}{± 0.62}} & \makecell[c]{15.90\\ \scalebox{\std}{± 0.57}} & \makecell[c]{17.90\\ \scalebox{\std}{± 1.04}} & \makecell[c]{13.12\\ \scalebox{\std}{± 0.78}} & \makecell[c]{8.46\\ \scalebox{\std}{± 0.35}} & \makecell[c]{12.62\\ \scalebox{\std}{± 0.33}} & \makecell[c]{21.53\\ \scalebox{\std}{± 1.26}} & \makecell[c]{16.35\\ \scalebox{\std}{± 0.68}} & \makecell[c]{22.57\\ \scalebox{\std}{± 1.26}} & \cellcolor[HTML]{EDEEFF}\makecell[c]{18.37\\ \scalebox{\std}{± 0.29}} \\
\end{tabularx}
\end{table}

\begin{table}[H]
\centering
\caption{Mean absolute error (MAE) (\%) of the accuracy estimation on fully CIFAR100-C. Averaged over three different random seeds.}
\label{tab:main_experiment_cifar100_full}
\scriptsize
\setlength\tabcolsep{3pt}
\begin{tabularx}{\linewidth}{ll*{16}Y}
\Xhline{2\arrayrulewidth}
 &  & \multicolumn{3}{c}{Noise} & \multicolumn{4}{c}{Blur} & \multicolumn{4}{c}{Weather} & \multicolumn{4}{c}{Digital} &  \\ 
\addlinespace[-0.05cm]
\cmidrule(lr){3-5} \cmidrule(lr){6-9} \cmidrule(lr){10-13} \cmidrule(lr){14-17}
\addlinespace[-0.05cm]
TTA Method & Acc. Estimation & Gau. & Shot & Imp. & Def. & Gla. & Mot. & Zoom & Snow & Fro. & Fog & Brit. & Cont. & Elas. & Pix. & JPEG & \cellcolor[HTML]{EDEEFF}Avg.($\downarrow$) \\ \midrule
  & SrcValid & \makecell[c]{46.38\\ \scalebox{\std}{± 1.17}} & \makecell[c]{45.01\\ \scalebox{\std}{± 1.16}} & \makecell[c]{51.81\\ \scalebox{\std}{± 1.79}} & \makecell[c]{31.42\\ \scalebox{\std}{± 0.67}} & \makecell[c]{49.43\\ \scalebox{\std}{± 0.58}} & \makecell[c]{33.51\\ \scalebox{\std}{± 0.64}} & \makecell[c]{31.37\\ \scalebox{\std}{± 0.35}} & \makecell[c]{39.05\\ \scalebox{\std}{± 0.33}} & \makecell[c]{38.89\\ \scalebox{\std}{± 0.28}} & \makecell[c]{34.25\\ \scalebox{\std}{± 0.38}} & \makecell[c]{28.72\\ \scalebox{\std}{± 0.33}} & \makecell[c]{33.04\\ \scalebox{\std}{± 0.61}} & \makecell[c]{41.81\\ \scalebox{\std}{± 0.75}} & \makecell[c]{35.52\\ \scalebox{\std}{± 0.51}} & \makecell[c]{44.24\\ \scalebox{\std}{± 0.66}} & \cellcolor[HTML]{EDEEFF}\makecell[c]{38.96\\ \scalebox{\std}{± 0.22}} \\
\end{tabularx}
\end{table}

\begin{table}[H]
\centering
\caption{Mean absolute error (MAE) (\%) of the accuracy estimation on fully ImageNet-C. Averaged over three different random seeds.}
\label{tab:main_experiment_imagenet_full}
\scriptsize
\setlength\tabcolsep{3pt}
\begin{tabularx}{\linewidth}{ll*{16}Y}
\Xhline{2\arrayrulewidth}
 &  & \multicolumn{3}{c}{Noise} & \multicolumn{4}{c}{Blur} & \multicolumn{4}{c}{Weather} & \multicolumn{4}{c}{Digital} &  \\ 
\addlinespace[-0.05cm]
\cmidrule(lr){3-5} \cmidrule(lr){6-9} \cmidrule(lr){10-13} \cmidrule(lr){14-17}
\addlinespace[-0.05cm]
TTA Method & Acc. Estimation & Gau. & Shot & Imp. & Def. & Gla. & Mot. & Zoom & Snow & Fro. & Fog & Brit. & Cont. & Elas. & Pix. & JPEG & \cellcolor[HTML]{EDEEFF}Avg.($\downarrow$) \\ \midrule
 & SrcValid & \makecell[c]{53.54\\ \scalebox{\std}{± 1.08}} & \makecell[c]{52.12\\ \scalebox{\std}{± 1.12}} & \makecell[c]{53.28\\ \scalebox{\std}{± 0.95}} & \makecell[c]{54.72\\ \scalebox{\std}{± 0.95}} & \makecell[c]{53.66\\ \scalebox{\std}{± 1.07}} & \makecell[c]{42.96\\ \scalebox{\std}{± 0.92}} & \makecell[c]{32.78\\ \scalebox{\std}{± 0.83}} & \makecell[c]{37.25\\ \scalebox{\std}{± 0.96}} & \makecell[c]{38.77\\ \scalebox{\std}{± 0.97}} & \makecell[c]{25.06\\ \scalebox{\std}{± 0.94}} & \makecell[c]{10.47\\ \scalebox{\std}{± 0.67}} & \makecell[c]{53.56\\ \scalebox{\std}{± 0.77}} & \makecell[c]{27.86\\ \scalebox{\std}{± 0.85}} & \makecell[c]{22.35\\ \scalebox{\std}{± 0.70}} & \makecell[c]{28.55\\ \scalebox{\std}{± 0.71}} & \cellcolor[HTML]{EDEEFF}\makecell[c]{39.13\\ \scalebox{\std}{± 0.89}} \\
\end{tabularx}
\end{table}

\begin{table}[H]
\centering
\caption{Mean absolute error (MAE) (\%) of the accuracy estimation on continual CIFAR10-C. Averaged over three different random seeds.}
\label{tab:main_experiment_cifar10_cont}
\scriptsize
\setlength\tabcolsep{3pt}
\begin{tabularx}{\linewidth}{ll*{16}Y}
\Xhline{2\arrayrulewidth}
& & \multicolumn{15}{l}{$t$ 
    \begin{tikzpicture}[overlay, remember picture]
      \draw[->, line width=0.2pt] (0,0.075) -- (12.75,0.075);
    \end{tikzpicture}
    } &\\
\addlinespace[-0.1cm]
TTA Method & Acc. Estimation & Gau. & Shot & Imp. & Def. & Gla. & Mot. & Zoom & Snow & Fro. & Fog & Brit. & Cont. & Elas. & Pix. & JPEG & \cellcolor[HTML]{EDEEFF}Avg.($\downarrow$) \\ 
\midrule
 & SrcValid & \makecell[c]{24.85\\ \scalebox{\std}{± 0.83}} & \makecell[c]{17.83\\ \scalebox{\std}{± 1.52}} & \makecell[c]{22.28\\ \scalebox{\std}{± 0.17}} & \makecell[c]{8.89\\ \scalebox{\std}{± 1.17}} & \makecell[c]{19.20\\ \scalebox{\std}{± 3.28}} & \makecell[c]{8.62\\ \scalebox{\std}{± 3.45}} & \makecell[c]{7.16\\ \scalebox{\std}{± 2.71}} & \makecell[c]{9.00\\ \scalebox{\std}{± 3.73}} & \makecell[c]{8.13\\ \scalebox{\std}{± 2.54}} & \makecell[c]{7.34\\ \scalebox{\std}{± 2.32}} & \makecell[c]{4.44\\ \scalebox{\std}{± 0.57}} & \makecell[c]{6.28\\ \scalebox{\std}{± 1.89}} & \makecell[c]{7.70\\ \scalebox{\std}{± 3.52}} & \makecell[c]{5.22\\ \scalebox{\std}{± 1.67}} & \makecell[c]{5.68\\ \scalebox{\std}{± 2.27}} & \cellcolor[HTML]{EDEEFF}\makecell[c]{10.84\\ \scalebox{\std}{± 1.83}} \\
\end{tabularx}
\end{table}

\begin{table}[H]
\centering
\caption{Mean absolute error (MAE) (\%) of the accuracy estimation on continual CIFAR100-C. Averaged over three different random seeds.}
\label{tab:main_experiment_cifar100_cont}
\scriptsize
\setlength\tabcolsep{3pt}
\begin{tabularx}{\linewidth}{ll*{16}Y}
\Xhline{2\arrayrulewidth}
& & \multicolumn{15}{l}{$t$ 
    \begin{tikzpicture}[overlay, remember picture]
      \draw[->, line width=0.2pt] (0,0.075) -- (12.75,0.075);
    \end{tikzpicture}
    } &\\
\addlinespace[-0.1cm]
TTA Method & Acc. Estimation & Gau. & Shot & Imp. & Def. & Gla. & Mot. & Zoom & Snow & Fro. & Fog & Brit. & Cont. & Elas. & Pix. & JPEG & \cellcolor[HTML]{EDEEFF}Avg.($\downarrow$) \\ \midrule
 & SrcValid & \makecell[c]{46.38\\ \scalebox{\std}{± 1.17}} & \makecell[c]{32.85\\ \scalebox{\std}{± 3.59}} & \makecell[c]{28.45\\ \scalebox{\std}{± 3.99}} & \makecell[c]{12.51\\ \scalebox{\std}{± 0.84}} & \makecell[c]{16.92\\ \scalebox{\std}{± 2.40}} & \makecell[c]{5.75\\ \scalebox{\std}{± 0.60}} & \makecell[c]{3.84\\ \scalebox{\std}{± 0.73}} & \makecell[c]{3.88\\ \scalebox{\std}{± 1.45}} & \makecell[c]{2.77\\ \scalebox{\std}{± 0.87}} & \makecell[c]{2.32\\ \scalebox{\std}{± 0.35}} & \makecell[c]{2.12\\ \scalebox{\std}{± 0.20}} & \makecell[c]{2.00\\ \scalebox{\std}{± 0.13}} & \makecell[c]{1.87\\ \scalebox{\std}{± 0.15}} & \makecell[c]{1.73\\ \scalebox{\std}{± 0.30}} & \makecell[c]{1.65\\ \scalebox{\std}{± 0.26}} & \cellcolor[HTML]{EDEEFF}\makecell[c]{11.00\\ \scalebox{\std}{± 0.58}} \\
\end{tabularx}
\end{table}

\begin{table}[H]
\centering
\caption{Mean absolute error (MAE) (\%) of the accuracy estimation on continual ImageNet-C. Averaged over three different random seeds.}
\label{tab:main_experiment_imagenet_cont}
\scriptsize
\setlength\tabcolsep{3pt}
\begin{tabularx}{\linewidth}{ll*{16}Y}
\Xhline{2\arrayrulewidth}
& & \multicolumn{15}{l}{$t$ 
    \begin{tikzpicture}[overlay, remember picture]
      \draw[->, line width=0.2pt] (0,0.075) -- (12.75,0.075);
    \end{tikzpicture}
    } &\\
\addlinespace[-0.1cm]
TTA Method & Acc. Estimation & Gau. & Shot & Imp. & Def. & Gla. & Mot. & Zoom & Snow & Fro. & Fog & Brit. & Cont. & Elas. & Pix. & JPEG & \cellcolor[HTML]{EDEEFF}Avg.($\downarrow$) \\ \midrule
 & SrcValid & \makecell[c]{53.54\\ \scalebox{\std}{± 1.08}} & \makecell[c]{48.82\\ \scalebox{\std}{± 1.21}} & \makecell[c]{47.13\\ \scalebox{\std}{± 1.16}} & \makecell[c]{48.84\\ \scalebox{\std}{± 1.07}} & \makecell[c]{44.94\\ \scalebox{\std}{± 1.16}} & \makecell[c]{34.83\\ \scalebox{\std}{± 1.25}} & \makecell[c]{26.16\\ \scalebox{\std}{± 1.18}} & \makecell[c]{32.41\\ \scalebox{\std}{± 1.21}} & \makecell[c]{31.80\\ \scalebox{\std}{± 1.02}} & \makecell[c]{21.17\\ \scalebox{\std}{± 1.05}} & \makecell[c]{8.93\\ \scalebox{\std}{± 0.73}} & \makecell[c]{43.25\\ \scalebox{\std}{± 0.72}} & \makecell[c]{20.57\\ \scalebox{\std}{± 0.66}} & \makecell[c]{17.02\\ \scalebox{\std}{± 0.49}} & \makecell[c]{20.10\\ \scalebox{\std}{± 0.27}} & \cellcolor[HTML]{EDEEFF}\makecell[c]{33.30\\ \scalebox{\std}{± 0.93}} \\
\end{tabularx}
\end{table}

\renewcommand{\arraystretch}{1.8}

\begin{table}[H]
\centering
\caption{Average accuracy improvement (\%p) with model recovery. Averaged over three different random seeds.}
\label{tab:model_recovery_all}
\scriptsize
\setlength\tabcolsep{3pt}
\begin{tabularx}{\linewidth}{ll*{16}Y}
\Xhline{2\arrayrulewidth}
& & \multicolumn{15}{l}{$t$ 
    \begin{tikzpicture}[overlay, remember picture]
      \draw[->, line width=0.2pt] (0,0.075) -- (12.6,0.075);
    \end{tikzpicture}
    } &\\
\addlinespace[-0.1cm]
 &  & Gau. & Shot & Imp. & Def. & Gla. & Mot. & Zoom & Snow & Fro. & Fog & Brit. & Cont. & Elas. & Pix. & JPEG & \cellcolor[HTML]{EDEEFF}Avg.($\uparrow$) \\ \midrule
 & Episodic~\cite{memo} & \makecell[c]{-10.41\\ \scalebox{\std}{± 0.85}} & \makecell[c]{-10.27\\ \scalebox{\std}{± 0.40}} & \makecell[c]{-4.58\\ \scalebox{\std}{± 0.93}} & \makecell[c]{22.41\\ \scalebox{\std}{± 1.13}} & \makecell[c]{14.47\\ \scalebox{\std}{± 0.72}} & \makecell[c]{38.62\\ \scalebox{\std}{± 2.00}} & \makecell[c]{44.44\\ \scalebox{\std}{± 2.82}} & \makecell[c]{41.68\\ \scalebox{\std}{± 2.66}} & \makecell[c]{44.55\\ \scalebox{\std}{± 3.61}} & \makecell[c]{51.44\\ \scalebox{\std}{± 2.92}} & \makecell[c]{62.95\\ \scalebox{\std}{± 1.48}} & \makecell[c]{61.34\\ \scalebox{\std}{± 1.23}} & \makecell[c]{50.74\\ \scalebox{\std}{± 1.07}} & \makecell[c]{53.72\\ \scalebox{\std}{± 0.45}} & \makecell[c]{42.53\\ \scalebox{\std}{± 0.57}} & \cellcolor[HTML]{EDEEFF}\makecell[c]{33.58\\ \scalebox{\std}{± 1.04}}
 \\
\end{tabularx}
\end{table}


\end{document}